\documentclass{article}

 \usepackage[main, final]{neurips_2026}

\usepackage[utf8]{inputenc} 
\usepackage[T1]{fontenc}    
\usepackage{nameref, hyperref} 
\usepackage{url}            
\usepackage{booktabs}       
\usepackage{amsfonts}       
\usepackage{nicefrac}       
\usepackage{microtype}      
\usepackage{xcolor}         
\usepackage{graphicx}          
\usepackage{wrapfig}           
\usepackage{amssymb}           
\usepackage{amsmath}           
\usepackage{bm}                
\usepackage{chngcntr}          

\title{Learning Reveals Invisible Structure \\ in Low-Rank RNNs}
\author{
\begin{tabular}{c@{\hskip 0.4in}c}
\textbf{Yoav Ger} & \textbf{Omri Barak} \\
\textnormal{Technion}\thanks{%
\begin{minipage}[t]{\linewidth}
\raggedright
Ruth and Bruce Rappaport Faculty of Medicine and Network Biology Research Laboratory\\
Technion -- Israel Institute of Technology
\end{minipage}} & \textnormal{Technion}\footnotemark[1] \\
\textnormal{\texttt{yoav.ger@campus.technion.ac.il}} & \textnormal{\texttt{omri.barak@gmail.com}} \\ 
\end{tabular}
}
\begin{document}
\maketitle

\begin{abstract}
    Learning in neural systems arises from synaptic changes that reshape the representations underlying behavior. While low-rank recurrent neural networks (RNNs) have emerged as a powerful framework for linking connectivity to function, a theoretical understanding of their learning process remains elusive. Here, we extend the low-rank framework from activity to learning by deriving gradient-descent dynamics directly in a reduced overlap space. We formulate a closed-form, low-dimensional system of ODEs that governs learning in this space, exact for linear RNNs and asymptotically exact for nonlinear RNNs in the large-$N$ Gaussian limit. Central to our analysis is a distinction between two classes of overlaps: \textit{loss-visible} overlaps, which fully determine network activity, output, and loss, and \textit{loss-invisible} overlaps, which do not affect function but are required to describe learning. We illustrate the consequences of this decomposition through two phenomena. First, we show that learning can serve as a perturbation that exposes differences in connectivity between functionally equivalent networks. Second, we show that loss-invisible overlaps can act as memory variables that encode training history, and characterize the conditions under which this occurs. Finally, we present several testable predictions for biological learning experiments derived from our theory. 
\end{abstract}

\section{Introduction}
Learning is a hallmark of intelligent systems, whether biological or artificial~\cite{barron2015embracing, hennig2021learning, saxe2021if}. In neuroscience, a central paradigm posits that learning arises from synaptic changes within neural circuits that reshape the internal dynamics (i.e., activity) and representations underlying behavior~\cite{thompson1986neurobiology, magee2020synaptic}. However, directly linking microscopic circuit-level plasticity to macroscopic behavioral outcomes remains a fundamental challenge~\cite{sadtler2014neural, humeau2019next}. One possible reason for this difficulty, at least in theory, lies in the disparity of scales~\cite{gao2015simplicity}. Adaptation occurs in a high-dimensional space of synaptic parameters (analogous to the overparameterized weight space in artificial neural networks), while the resulting functions or behaviors are much lower-dimensional, often by orders of magnitude. This mismatch renders the mapping from function to connectivity intrinsically ill-posed~\cite{das2020systematic}, raising fundamental questions about degeneracy~\cite{edelman2001degeneracy, prinz2004similar, albantakis2024brain} and identifiability~\cite{albertini1993neural,roeder2021linear,braun2025not} in neural systems.

A promising framework for addressing these challenges is low-rank recurrent neural networks. In these models, recurrent connectivity is constrained to be low-rank, such that the effective mapping from connectivity to network dynamics and function is fully described by a small set of macroscopic overlap variables~\cite{mastrogiuseppe2018linking}. This reduction has made low-rank RNNs a powerful model for studying recurrent computation, including an analysis of the networks' dynamical properties~\cite {schuessler2020dynamics, beiran2021shaping}, the design of engineered networks that implement prescribed computations~\cite{hopfield1982neural, eliasmith2003neural, marschall2025theory}, and work showing how low-rank structure emerges through training~\cite{schuessler2020interplay,dubreuil2022role,valente2022extracting}. While recent work has begun to analyze the learning dynamics of RNNs~\cite{bordelon2025dynamically, proca2025learning}, these approaches have largely been developed outside the low-rank framework. Consequently, it remains unclear whether the overlap view—so successful in describing network function—can be extended to account for learning while retaining a similar low-dimensional description.

To bridge this gap, we extend the low-rank framework from network activity to learning dynamics. By expressing gradient descent updates directly in terms of scalar overlaps, we obtain a closed-form, low-dimensional description of learning. This derivation reveals that the resulting dynamics are not equivalent to naive gradient descent in overlap space, but are rather shaped by a preconditioning metric that captures the geometry of the high-dimensional parameter space. Interestingly, for low-rank RNNs, this metric can be computed explicitly and depends on additional overlaps beyond those that determine the current function, thereby revealing structural constraints on learning that are invisible at the level of function alone.

Our \textbf{contributions} can be summarized as follows:
\begin{itemize}
    \item \textbf{Technical:} We extend the low-rank framework to learning by deriving a closed-form system of ODEs for the overlap dynamics in low-rank RNNs. These are exact in the linear case and asymptotically exact in the Gaussian nonlinear case as $N \to \infty$, providing, to the best of our knowledge, the first analytical description of learning in nonlinear task-trained RNNs.
    
    \item \textbf{Conceptual:} A key consequence of our technical derivation is a partition of connectivity into two groups: \textit{loss-visible} overlaps, which fully determine the network’s activity, output, and loss, and \textit{loss-invisible} overlaps, which are functionally silent yet shape the trajectory of learning. In our setting, the boundary between these groups is determined by the activation function (linear/nonlinear), although it may generally also depend on the architecture, input--output map, and task.
    
    \item \textbf{Implications:} We illustrate the implications of this partition through two central phenomena of neural learning: (i) degeneracy -- networks with identical function can have distinct connectivity, with learning resolving this ambiguity. Thus, observing how a system learns can serve as a non-invasive probe of underlying structure, an idea we term \emph{perturbation-by-learning}. (ii) memory -- \textit{loss-invisible} overlaps can serve as memory variables, encoding aspects of past training history without affecting network function. We show that memory is generally unreliable in linear networks, with its presence depending on the learning rule, while in nonlinear networks it emerges more readily.    
\end{itemize}

\section{Preliminaries}
We study a high-dimensional RNN trained via gradient descent. While our framework is general, we develop it here for RNNs, a canonical model in theoretical neuroscience~\cite{hopfield1982neural,sompolinsky1988chaos, barak2017recurrent}. Throughout, bold lowercase letters denote vectors (e.g., $\bm{z}$), bold uppercase letters denote matrices (e.g., $\bm{W}$), and plain symbols denote scalars. For two vectors $\bm{u}, \bm{v} \in \mathbb{R}^N$, we define their scaled overlap by $\sigma_{vu} = \tfrac{1}{N}\bm{v}^\top \bm{u}$ and the squared norm $\|\bm{v}\|^2 = \tfrac{1}{N}\bm{v}^\top \bm{v}$, so both remain $O(1)$ as $N \to \infty$. Within-episode (trial) time is indexed by $t$, and learning time (across episodes) by $\tau$. Gradients with respect to parameters $\bm{\theta}$ are written $\nabla_{\bm{\theta}}$, and $\dot{\bm{\theta}} = d\bm{\theta}/d\tau$ denotes differentiation with respect to learning time.

\begin{figure}[!t]
    \centering
    \includegraphics[width=\linewidth]{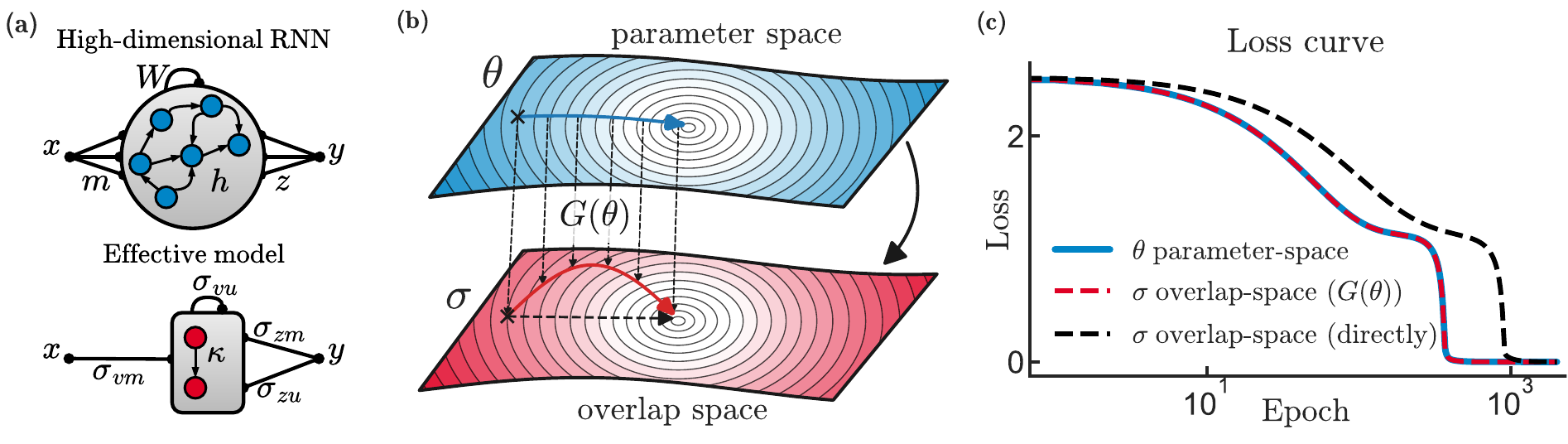}
\caption{\textbf{(a)} High-dimensional RNN in parameter $\bm{\theta}$-space (top): input $x$ drives activity $\bm h$ through input weights $\bm m$, recurrent connectivity $\bm W$, and readout weights $\bm z$ to produce output $y$. For a low-rank RNN, the same input–output function is captured by an effective model described by a small set of scalar overlaps $\bm{\sigma}$ (bottom). \textbf{(b)} Schematic illustration of a learning trajectory in the loss landscape over the parameter $\bm \theta$ (top, blue arrow) and its projection onto the loss landscape over the overlap $\bm \sigma$ via $\bm G(\bm \theta)$ (bottom, red arrow), highlighting how this mapping can alter the perceived trajectory. Crucially, the projected dynamics can differ from direct optimization in overlap space (black dashed arrow), reflecting structural constraints of the parameterization. \textbf{(c)} Concrete example of (b), showing the loss dynamics of training a low-rank linear RNN on a filter task. Optimization in parameter space (blue) and the corresponding overlap dynamics induced by $\bm G$ (red dashed) closely agree, whereas direct optimization in overlap space (black dashed) produces different dynamics.}
\label{fig:1}
\end{figure}

\paragraph{RNN model}
We consider a rate-based RNN with $N$ neurons (Fig.~\ref{fig:1}a, top). Its continuous-time dynamics and readout are
\begin{equation}
    \dot{\bm{h}}(t) = -\bm{h}(t) + \frac{1}{\sqrt{N}}\,\bm{W}\,\phi\!\left(\bm{h}(t)\right) + \bm{m}\,x(t), 
    \qquad 
    \hat{y}(t) = \frac{1}{N}\,\bm{z}^\top \phi\!\left(\bm{h}(t)\right)
    \label{Eq.1}
\end{equation}
where $\bm{h}(t) \in \mathbb{R}^N$ is the hidden state, $\phi(\cdot)$ is an element-wise activation function, and $\bm{W} \in \mathbb{R}^{N \times N}$ is the recurrent connectivity matrix. For simplicity, we focus on a single-input, single-output network (extensions to multiple inputs/outputs are straightforward). The scalar input $x(t)$ enters through $\bm{m} \in \mathbb{R}^N$, and the scalar output $\hat{y}(t)$ is obtained via a linear readout with weights $\bm{z} \in \mathbb{R}^N$.

\paragraph{Learning setup}
The trainable parameters are collected as $\bm{\theta} = \{\bm{m}, \bm{W}, \bm{z}\}$ and updated across episodes to minimize the squared-error loss relative to a target $y^\star(t)$
\begin{equation}
    \mathcal{L}
    =
    \int_{0}^{T} \bigl[\hat{y}(t) - y^\star(t)\bigr]^2\,dt
    \label{Eq.2}
\end{equation}
where $T$ denotes the total episode duration. We analyze learning in the gradient-flow limit $\eta \to 0$, where parameter updates follow
\begin{equation}
    \dot{\bm{\theta}} = -\nabla_{\bm{\theta}}\mathcal{L}
    \label{Eq.3}
\end{equation}
While this formulation provides an exact description of learning in the full parameter space, the resulting dynamics are high-dimensional and difficult to interpret. The key observation we exploit is that, in many cases of interest, the network output—and therefore the loss—depend on $\bm{\theta}$ only through a reduced set of variables (i.e., redundancy). Thus, although learning occurs in a high-dimensional space, behavior evolves along far fewer effective degrees of freedom. To make this point concrete, we next specialize to low-rank RNNs and derive a reduced description of learning. 

\section{Low-rank linear RNN}
To build intuition, we begin with a simple tractable setting: a rank-1 RNN with linear activation $\phi(\cdot)=\mathrm{id}$. In this case, the dynamics of Eq.~\eqref{Eq.1} simplify to
\begin{equation}
\dot{\bm h}(t) = -\bm h(t) + \frac{1}{N}\bm u \bm v^\top \bm h(t) + \bm mx(t)
\label{Eq.4}
\end{equation}
where $\bm u, \bm v \in \mathbb{R}^{N}$ define the rank-1 recurrent connectivity. The trainable parameters are four vectors in $\mathbb{R}^N$, $\bm{\theta}=\{\bm m,\bm u,\bm v,\bm z\}$, corresponding to the input, left and right recurrent, and readout vectors. Assuming a zero initial condition \(\bm h(0)=\bm 0\), known results on low-rank RNNs~\cite{mastrogiuseppe2018linking} imply that the hidden-state dynamics are confined to the two-dimensional subspace \(\mathrm{span}\{\bm m,\bm u\}\), and can therefore be written as
\begin{equation}
    \bm{h}(t)
    = \kappa_m(t)\,\bm{m} + \kappa_u(t)\,\bm{u},
    \qquad
    \boldsymbol{\kappa}(t) =
    \begin{bmatrix}
        \kappa_m(t) \\
        \kappa_u(t)
    \end{bmatrix} \in \mathbb{R}^2
\label{Eq.5}
\end{equation}
where $\kappa_m(t)$ and $\kappa_u(t)$ denote the coordinates of the activity along the input and recurrent modes. Substituting this ansatz into Eq.~\eqref{Eq.4} yields a 2D effective RNN (Fig.\ref{fig:1}a, bottom) with activity dynamics and output 
\begin{equation}
    \dot{\boldsymbol{\kappa}}(t)
    = -\boldsymbol{\kappa}(t) +
    \begin{bmatrix}
        0 & 0\\
        \tfrac{1}{N}\bm{v}^{\!\top}\bm{m} & \tfrac{1}{N}\bm{v}^{\!\top}\bm{u}
    \end{bmatrix}
    \boldsymbol{\kappa}(t)
    +
    \begin{bmatrix}
        1 \\[2pt] 0
    \end{bmatrix}
    x(t),
    \qquad
    \hat{y}(t) =
    \frac{1}{N}
    \begin{bmatrix}
        \bm{z}^{\!\top}\bm{m} &
        \bm{z}^{\!\top}\bm{u}
    \end{bmatrix}
    \boldsymbol{\kappa}(t)
\label{Eq.6}
\end{equation}
Thus, the $N$-dimensional RNN reduces exactly to a 2D system whose input--output behavior is fully determined by four scalar overlaps
\begin{equation}
\sigma_{zm}=\tfrac{1}{N}\bm{z}^{\!\top}\bm{m},\quad
\sigma_{zu}=\tfrac{1}{N}\bm{z}^{\!\top}\bm{u},\quad
\sigma_{vm}=\tfrac{1}{N}\bm{v}^{\!\top}\bm{m},\quad
\sigma_{vu}=\tfrac{1}{N}\bm{v}^{\!\top}\bm{u}
\label{Eq.7}
\end{equation}
We collect these quantities into the vector $\bm{\sigma}=(\sigma_{zm},\sigma_{zu},\sigma_{vm},\sigma_{vu})$, and refer to them as the \emph{loss-visible} overlaps, since they fully determine the within-episode dynamics and thus the loss. Crucially, although the loss depends only on $\bm{\sigma}$, optimization is performed in the high-dimensional parameter space $\bm{\theta}$. As a result, the trajectory of the overlaps induced by learning can differ from the one obtained by directly minimizing $\bm{\sigma}$ (Fig.~\ref{fig:1}b). To connect parameter-space learning with the induced dynamics in overlap space, we consider the Jacobian of $\bm{\sigma}$ with respect to $\bm{\theta}$
\begin{equation}
\bm D(\bm\theta) = \frac{\partial \bm\sigma}{\partial \bm\theta} \in \mathbb R^{4\times 4N}
\label{Eq.8}
\end{equation}
Because the loss depends on the parameters only through the overlaps, the chain rule gives $\nabla_{\bm{\theta}}\mathcal{L}=\bm D(\bm\theta)^{\top}\nabla_{\bm{\sigma}}\mathcal{L}$, where $\nabla_{\bm\sigma}\mathcal L \in \mathbb R^{4}$. Under gradient flow in parameter space, the overlaps evolve as
\begin{equation}
\dot{\bm{\sigma}}
= \bm D(\bm\theta)\,\dot{\bm{\theta}} = - \bm D(\bm\theta)\bm D(\bm\theta)^{\top}\nabla_{\bm{\sigma}}\mathcal{L}
\label{Eq.9}
\end{equation}
where $\bm G(\bm\theta)=\bm D(\bm\theta)\bm D(\bm\theta)^{\!\top}$ is a symmetric, positive semi-definite Gram matrix that defines the effective learning metric on overlap space. Thus, $\bm G(\bm\theta)$ acts as a preconditioner, reshaping $\nabla_{\bm\sigma}\mathcal L$ according to the geometry inherited from parameter space. Interestingly, for a rank-1 RNN, the matrix $\bm G(\bm\theta)$ can be computed in closed form (App.~\ref{app:A3}) and is given by
\begin{equation}
\bm{G}(\bm\theta) = \frac{1}{N}
\begin{bmatrix}
\|\bm{m}\|^2+\|\bm{z}\|^2 & \sigma_{mu} & \sigma_{zv} & 0\\
\sigma_{mu} & \|\bm{u}\|^2+\|\bm{z}\|^2 & 0 & \sigma_{zv}\\
\sigma_{zv} & 0 & \|\bm{m}\|^2+\|\bm{v}\|^2 & \sigma_{mu}\\
0 & \sigma_{zv} & \sigma_{mu} & \|\bm{u}\|^2+\|\bm{v}\|^2
\end{bmatrix}
\label{Eq.10}
\end{equation}
A direct inspection of $\bm G(\bm\theta)$ shows six additional quantities beyond the loss-visible overlaps $\bm\sigma$. These include other overlaps ($\sigma_{mu}\,,\sigma_{zv}$) as well as all the squared norms of the parameter vectors, none of which contribute to the loss. We group them into
\begin{equation}
\tilde{\bm\sigma} = (\sigma_{mu}, \sigma_{zv}, \|\bm m\|^2, \|\bm u\|^2, \|\bm v\|^2, \|\bm z\|^2)
\label{Eq.11}
\end{equation}
and refer to them as \emph{loss-invisible} overlaps. Together, $(\bm\sigma, \tilde{\bm\sigma})$ provide a closed, 10-dimensional description of the learning dynamics in overlap space. In other words, initializations with the same ten overlaps yield identical input--output functions and learning trajectories, despite potentially differing along the high-dimensional parameter space. Note that the loss-invisible overlaps evolve analogously as the loss-visible ones, with their dynamics (i.e., $\dot{\tilde{\bm\sigma}}$) derived explicitly in App.~\ref{app:A2}. 

To verify that this compact low-dimensional description faithfully captures high-dimensional learning, we train a rank-1 linear RNN on a simple filter task. In this task, the network is trained to emulate the output of a first-order exponential filter driven by white noise input~\cite{bordelon2025dynamically} (see App.~\ref{app:A4} for full details). The loss obtained from numerical simulation of gradient descent in the full parameter space matches exactly the prediction obtained by integrating our 10D ODE system in overlap space (Fig.~\ref{fig:1}c; and also see Fig.~\ref{fig:5}). In contrast, directly optimizing $\mathcal L(\bm{\sigma})$ produces qualitatively different loss dynamics (App.~\ref{app:A5} for a detailed comparison between optimization in both spaces). A complete derivation of this section is provided in App.~\ref{app:A}.

Before proceeding, we highlight an important point. The matrix $\bm{G}$ is a sub-block of a larger matrix that arises when considering the full set of ten quadratic overlaps among the four vectors $\bm \theta=\{\bm{m}, \bm{u}, \bm{v}, \bm{z}\}$. Differentiating this set with respect to $\bm{\theta}$ yields an augmented Jacobian $\bar{\bm D}(\bm \theta) \in \mathbb{R}^{10 \times 4N}$ and a corresponding $10 \times 10$ Gram matrix $\bar{\bm G}(\bm \theta)$ that jointly governs the evolution of both visible and invisible overlaps (see App.~\ref{app:C} and also App.~\ref{app:D} for a more general and compact derivation).

\section{Implications of visible and invisible overlaps}
We now examine two consequences of decomposing connectivity into loss-visible and loss-invisible overlaps in the context of learning.

\begin{figure}[!t]
    \centering
    \includegraphics[width=\linewidth]{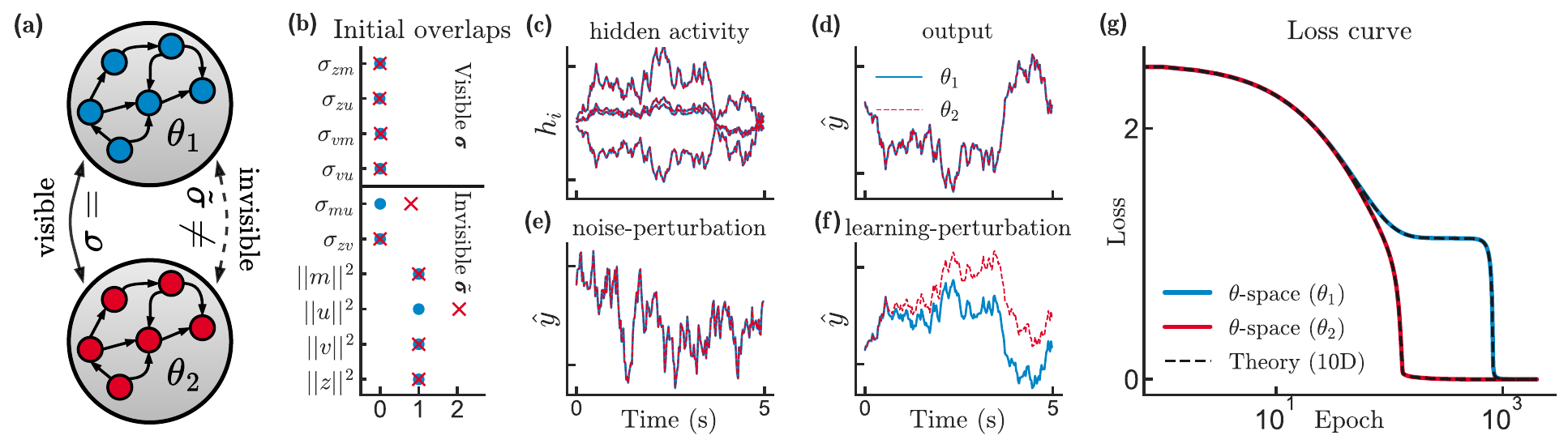}
\caption{\textbf{(a)} Two RNNs parametrized with $\bm \theta_1$ and $\bm \theta_2$ share identical loss-visible $\bm\sigma$ ($=$) but differ in loss-invisible $\tilde{\bm\sigma}$ ($\neq$) overlaps. \textbf{(b)} Initial values of all overlaps (blue: $\bm \theta_1$, red: $\bm \theta_2$) show identical visible components (top) but differences in the invisible components (bottom), including $\sigma_{mu}$ and $||u||^2$. \textbf{(c–d)} Because the input–output function depends only on the visible set, their hidden activity (c) and outputs (d) are indistinguishable. \textbf{(e)} Input noise perturbations likewise fail to differentiate the networks. \textbf{(f)} Once learning is turned on, differences in connectivity are revealed in the recorded outputs. \textbf{(g)} These invisible differences also emerge in the learning trajectories: $\bm \theta_1$ exhibits a transient plateau (blue) absent in $\bm \theta_2$ (red). The theory accurately captures both learning dynamics.}
\label{fig:2}
\end{figure}

\subsection{Learning reveals hidden degeneracy}
A direct consequence of this framework is the separation between two sets of overlaps. The first, loss-visible overlaps, determine the within-episode dynamics, output, and loss. The second, loss-invisible overlaps, leave the input--output function unchanged but shape the effective learning metric $\bm G(\bm\theta)$. As a result, two networks can implement the same function yet differ in their underlying connectivity structures. Although such degeneracies are well documented in modern machine learning theory~\cite{braun2025not,huang2025measuring}, our framework provides a direct characterization of these hidden degrees of freedom and shows how learning can reveal them. In this sense, learning acts as a \emph{perturbation} that reveals otherwise hidden degeneracies in the connectivity.

To illustrate this effect, consider two rank-1 linear RNNs parameterized by $\bm\theta_1$ and $\bm\theta_2$ (Fig.~\ref{fig:2}a), constructed to share identical loss-visible overlaps while differing in their loss-invisible overlaps (Fig.~\ref{fig:2}b). Because the visible overlaps coincide, the two networks exhibit identical hidden dynamics and outputs (Figs.~\ref{fig:2}c,d), and remain indistinguishable even under input noise-perturbations (Fig.~\ref{fig:2}e). However, once learning is initiated (training on the filter task), the difference in loss-invisible overlaps leads to different parameter updates, causing the two networks to produce distinct outputs in response to the same input (Fig.~\ref{fig:2}f). This divergence is also reflected in the loss dynamics (Fig.~\ref{fig:2}g), where network-1 exhibits a pronounced plateau that is absent in network-2. Importantly, both networks's learning dynamics are fully captured by our reduced 10D theory (dashed black).

\subsection{Memory and its absence in invisible overlaps}
\begin{figure}[!t]
    \centering
    \includegraphics[width=\linewidth]{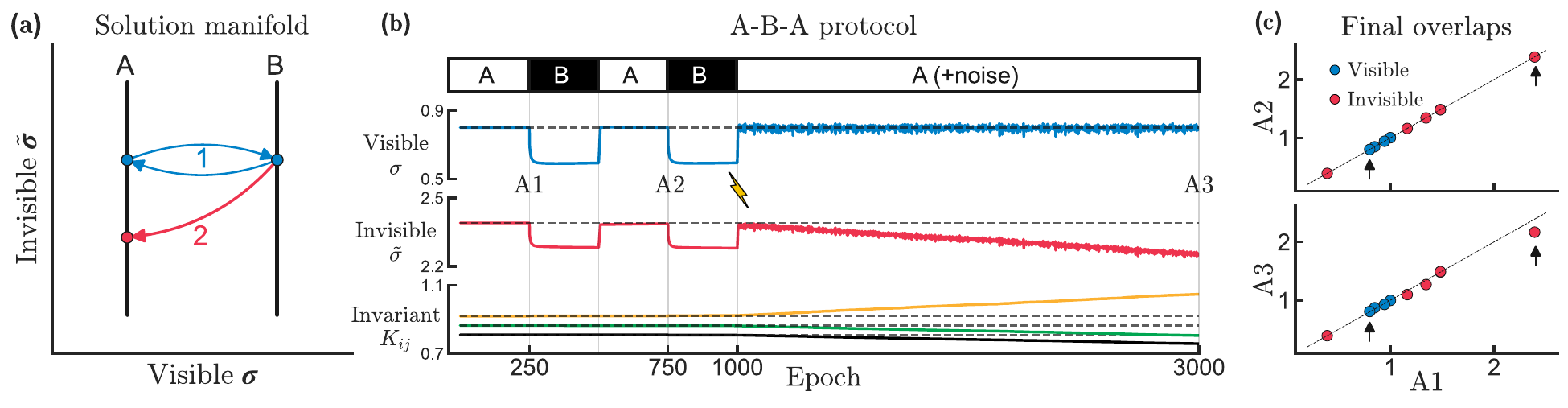}
\caption{\textbf{(a)} Illustration of the solution manifold in overlap space for tasks A and B. For each task, the loss fixes a subset (or all) of the loss-visible overlaps $\bm{\sigma}$ (x-axis), leaving a continuous manifold (black lines) of equivalent solutions parameterized by \textit{all} the loss-invisible overlaps $\tilde{\bm{\sigma}}$ (y-axis). Under an A–B–A training protocol, retraining on task~A can either (1) recover the original solution (blue) or (2) converge to a different, history-dependent solution on the manifold (red). The panel shows a two-dimensional schematic; in general, the visible and invisible sets contain multiple overlaps. \textbf{(b)} Learning trajectories during the A–B–A protocol. Top: representative loss-visible overlap ($\sigma_{vu}$, blue). Middle: representative loss-invisible overlap ($\|\bm u\|^2$, red). Bottom: three sampled entries of the conserved matrix $\bm K$. Under vanilla gradient descent, retraining on task~A restores both visible and invisible overlaps (compare A2, epoch 750 with A1, epoch 250), while $\bm K$ remains constant. Introducing label noise at epoch 1000 breaks the conservation of $\bm K$, inducing a directed drift in the invisible overlap (middle red) while leaving the visible overlap (top blue) unchanged. \textbf{(c)} Comparison of overlaps across task A solutions (A1, A2 and A3). Top: overlaps at A1 (epoch 250; x-axis) versus A2 (epoch 750; y-axis) lie on the identity line, demonstrating exact recovery. Bottom: with label noise A3 (epoch 3000; y-axis), invisible overlaps (red) deviate from the identity, whereas visible overlaps (blue) remain mostly unchanged. Arrows mark the overlaps depicted in (b).}
\label{fig:3}
\end{figure}

Turning our attention to the loss-invisible overlaps. Although they do not affect the network’s function, we showed above that they influence \textit{future} learning. This raises the question of whether they can also retain information about the history of \textit{past} learning, that is, serve as memory variables. To investigate this, we employ an A–B–A training protocol~\cite{confavreux2025memory}, in which the RNN is trained sequentially on task A, then task B, and finally retrained on task A. Because the loss constrains only the loss-visible overlaps, each task admits a continuous manifold of equivalent solutions parameterized by the loss-invisible directions (Fig.~\ref{fig:3}a). Thus, upon returning to task A, learning could either recover the original solution (blue) or find a different solution on the same manifold (red), revealing history dependence.

To determine which scenario is realized, we train a rank-1 linear RNN on a sequential filter task with two interleaved decay rates (Fig.~\ref{fig:3}b). Surprisingly, under vanilla gradient descent, we observe complete recovery: when retraining to task A, not only the loss-visible overlaps (expected), but also the loss-invisible overlaps (unexpectedly) return exactly to their original values (Fig.~\ref{fig:3}b, epoch 750 and Fig.~\ref{fig:3}c, top). This result indicates that the loss-invisible overlaps are not shaped by the training history, but are instead constrained by the task objective and the initialization. Indeed, our low-rank linear RNN falls within a class of matrix-factorized models~\cite{du2018algorithmic}, where the loss depends on the parameters only through a bilinear form involving two disjoint parameter matrices
\begin{equation}
\mathcal{L}(\bm{\theta}) = \mathcal{L}\left( 
\frac{1}{N}\begin{bmatrix}
\bm{z}^\top \\ \bm{v}^\top
\end{bmatrix}
\begin{bmatrix}
\bm{m} & \bm{u}
\end{bmatrix}
\right) = \mathcal{L}\left( 
\begin{pmatrix}
\sigma_{zm} & \sigma_{zu} \\
\sigma_{vm} & \sigma_{vu}
\end{pmatrix}
\right)
\label{Eq.12}
\end{equation}
For such models, gradient flow admits exact invariants of the learning dynamics (see App.~\ref{app:A6} for full derivation). In particular, the matrix
\begin{equation}
\bm K =\begin{bmatrix}
\bm z & \bm v
\end{bmatrix}
\begin{bmatrix}
\bm z^\top \\ \bm v^\top
\end{bmatrix}
-
\begin{bmatrix}
\bm m & \bm u
\end{bmatrix}
\begin{bmatrix}
\bm m^\top \\ \bm u^\top
\end{bmatrix}=
\bm z\bm z^\top
+\bm v\bm v^\top
-\bm m\bm m^\top
-\bm u\bm u^\top
\in \mathbb R^{N\times N} 
\label{Eq.13}
\end{equation}
is conserved throughout learning~\cite{saxe2013exact} (Fig.~\ref{fig:3}b, bottom). Consequently, training trajectories are confined to invariant manifolds set by the initialization. Thus, if the visible overlaps return to their original values, so will the invisible ones. This analysis implies that encoding memory in the loss-invisible overlaps requires breaking this invariant, which can be achieved either by modifying the architecture (see nonlinear below) or by altering the learning rule. In the latter case, adding label noise (which deviates from pure gradient flow) breaks the conservation of $\bm K$ (Fig.~\ref{fig:3}b, bottom, from epoch 1000 onward), causing its entries to drift. Notably, this perturbation induces a directed drift within the loss-invisible subspace while leaving loss-visible overlaps unchanged (Figs.~\ref{fig:3}b,c). Such behavior is consistent with SGD dynamics, where noise drives solutions toward flatter or lower-norm regions of the solution manifold~\cite{blanc2020implicit,ratzon2024representational}. In our setting, this corresponds to a reduction in loss-invisible quantities such as $\|\bm u\|^2$. In the Appendix, we present full trajectories for all ten overlaps (Fig.~\ref{fig:7}) and show that adaptive optimizers, such as Adam~\cite{kingma2014adam}, similarly break this invariant (Fig.~\ref{fig:8}).

\section{Low-rank nonlinear RNN}
\label{sec:5_nonlinear}
The results presented thus far apply to linear RNNs. We now extend the analysis to a nonlinear network, still within the rank-1 setting. Specifically, we consider a network with dynamics
\begin{equation}
\dot{\bm h}(t) = -\bm h(t) + \frac{1}{N}\bm u \bm v^\top \phi\bigl(\bm h(t)\bigr) + \bm m x(t),
\quad 
\phi(\bm h)=\mathrm{erf}\!\left(\tfrac{\sqrt{\pi}}{2}\bm h\right)
\end{equation}
Here, the nonlinear activation is the error function chosen for analytical tractability~\cite{marschall2025theory}, and the prefactor $\sqrt{\pi}/2$ ensures unit slope at the origin. Otherwise, the model is identical to the linear rank-1 RNN. To analyze the nonlinear case, we consider the limit $N\to\infty$ and assume that the components of the parameter vectors $\bm \theta$ are jointly Gaussian, following standard dynamical mean-field theory (DMFT)~\cite{mastrogiuseppe2018linking, sompolinsky1988chaos}. As in the linear case, the hidden state remains confined to the two-dimensional subspace spanned by $\bm m$ and $\bm u$, and can therefore be written as in Eq.~\eqref{Eq.5}. The key difference is that the recurrent input now depends nonlinearly on the state. Using Stein's Lemma, one obtains
\begin{equation}
\frac{1}{N}\bm v^\top\phi(\bm h(t)) =
\bigl(\sigma_{vm}\kappa_m(t)+\sigma_{vu}\kappa_u(t)\bigr)\,\langle \phi' \rangle
\end{equation}

For the erf nonlinearity, the average gain $\langle \phi' \rangle$ is analytically tractable and given by
\begin{equation}
    \begin{aligned}
        \langle \phi' \rangle &= \mathbb{E}_{g\sim\mathcal{N}(0,\Delta(t))}\!\left[\phi'(g)\right]= \left(1+\tfrac{\pi}{2}\Delta(t)\right)^{-1/2} \\[4pt]
        \Delta(t) &= \|\bm m\|^2 \kappa_m(t)^2+\|\bm u\|^2 \kappa_u(t)^2+2\,\sigma_{mu}\,\kappa_m(t)\kappa_u(t)
    \end{aligned}
    \label{Eq.16}
\end{equation}
Thus, under the Gaussian assumption, the input--output behavior of the nonlinear RNN is described by a finite set of macroscopic overlaps. Crucially, unlike in the linear case, the overlap $\sigma_{mu}$ as well as $\|\bm m\|^2$ and $\|\bm u\|^2$ now enter the dynamics through $\Delta(t)$ in Eq.~\eqref{Eq.16}, and therefore become \emph{loss-visible}, altering the previous visible--invisible separation. Accordingly, the loss no longer admits a disjoint bilinear dependence on the parameters as in Eq.~\eqref{Eq.12}, with important consequences described next. A complete mean-field derivation and full learning equation are provided in App.~\ref{app:B}. 

\begin{figure}[!t]
    \centering
    \includegraphics[width=\linewidth]{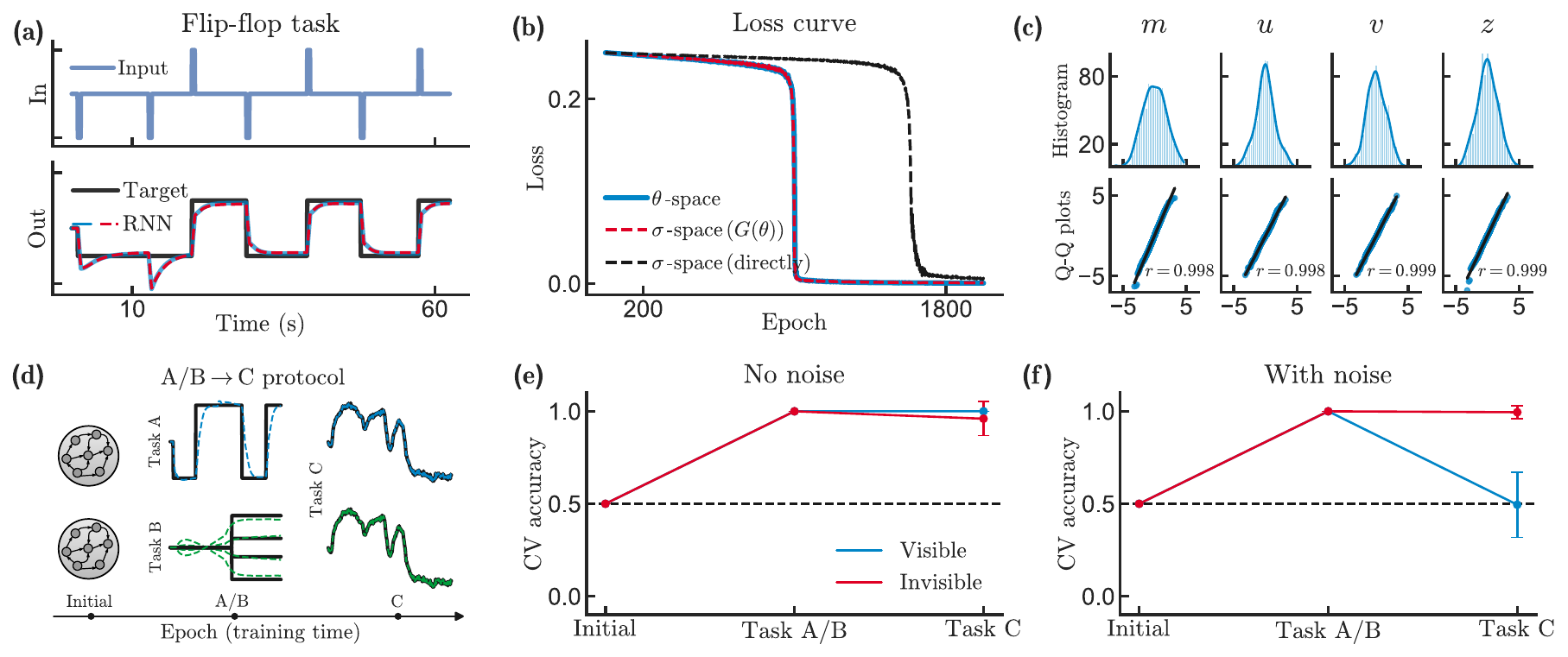}
\caption{\textbf{(a)} Example input sequence of the flip-flop task (top). Target output (black), high-dimensional RNN prediction (blue), and effective RNN prediction (red dashed) show excellent agreement (bottom). \textbf{(b)} Training loss of RNN in parameter space $\bm \theta$ (blue) and 10D overlap dynamics using $\bm{G}(\bm{\theta})$ (red dashed), which closely match, while direct optimization in $\bm{\sigma}$ space (black dashed), leads to different dynamics. \textbf{(c)} Distributions of parameter vector components $\bm{\theta}$ at convergence (top), along with corresponding Q--Q plots (bottom), showing that the Gaussian structure is preserved (correlation coefficient $r \geq 0.998$). \textbf{(d)} Schematic of the history-dependent training protocol ($\text{A/B} \rightarrow \text{C}$), where two identically initialized twin networks are first trained on either task A (flip-flop) or task B (stimulus integration), and then both on the same task C (emulating a teacher network). Solid black lines indicate targets, and dashed colored lines are the network outputs. \textbf{(e–f)} Cross-validation accuracy for decoding training history using loss-visible (blue) and loss-invisible (red) overlaps (error bars $\pm1$ s.d). (e) Without noise, decoding is possible from either set due to infinitesimal differences. (f) With small noise added to the overlaps ($\mathcal{N}(0, 0.1)$), only loss-invisible overlaps remain informative, while loss-visible overlaps drop to chance level.}
\label{fig:4}
\end{figure}

\subsection{Flip-flop task} 
To validate our nonlinear derivation, we consider the 1-bit flip-flop task (see App.~\ref{app:B4} for full details), a widely studied task in theoretical neuroscience~\cite{sussillo2013opening}. The task requires the network to maintain a stable internal state and update it only in response to brief, signed input pulses (Fig.~\ref{fig:4}a). As this requires bistability, which is absent in linear systems, it provides a natural setting to assess our nonlinear network. Training a high-dimensional nonlinear RNN on this task and comparing it with the corresponding 10D ODE theory, we observe close agreement in the loss curves (Fig.~\ref{fig:4}b). Furthermore, as in the linear case, directly optimizing in overlap space yields qualitatively different learning dynamics, underscoring the role of the structural preconditioning $\bm {G}(\bm{\theta})$. However, unlike the linear case, the validity of the theory relies on the components of the weight vectors $\bm{\theta}$ remaining approximately Gaussian. While gradient descent dynamics alone do not guarantee this condition, we numerically find that it holds throughout training for sufficiently small learning rates (Fig.~\ref{fig:4}c). In the Appendix, we test the limits of the Gaussian assumption, showing that training with larger learning rates or using alternative optimizers (e.g., Adam) leads to deviations from Gaussianity and discrepancies between theory and simulation (see App.~\ref{app:B41} and Fig.~\ref{fig:9}).

\subsection{Memory is encoded in the invisible overlaps}
Finally, we revisit whether the invisible overlaps can serve as memory variables encoding past training history. In linear networks trained with vanishingly small learning rate, the invisible overlaps are constrained by an invariant $\bm{K}$ and thus return to their original values. In contrast, in nonlinear networks, the altered visible--invisible separation breaks the invariant, opening the possibility for memory storage in the invisible set. To demonstrate this,  we devise a hypothetical training protocol where identically initialized networks (twin networks) undergo history-dependent training. Each network is first trained on a different task (A/B), then both on the same task (C), yielding identical outputs. We then ask whether training history can be decoded from the overlaps. We conjecture that, since the loss-visible overlaps determine the output, they must converge to the same values (up to task degeneracies) and are thus weakly informative. In contrast, loss-invisible overlaps do not affect the output and can retain distinct values, enabling them to encode the training history.

To test this, we train 20 networks (10 twin pairs) using the 10D overlap ODEs with the preconditioned metric $\bm G$, ensuring the Gaussian assumption is satisfied by construction while also affording a computational speedup. Each network is first trained on either Task A (flip-flop) or Task B (stimulus integration), and then on a common Task C, where it emulates the response of a teacher network (Fig.~\ref{fig:4}d; see also App.~\ref{app:B5} for full task details). We then train a classifier (logistic regression; similar results are obtained with other classifiers) to predict the training history (A or B) using either loss-visible or loss-invisible overlaps. Performance is evaluated via cross-validation accuracy over 50 random train–test splits and at three training checkpoints (Initial, Task A/B, and Task C). Our analysis reveals that, while both loss-visible and loss-invisible overlaps appear to be informative of the training history in the noiseless setting (Fig.~\ref{fig:4}e), this reflects infinitesimal differences arising from imperfect convergence (see Fig.~\ref{fig:11} for full overlap trajectories). However, consistent with our prediction, adding a small amount of noise to the overlaps (to mimic realistic conditions) reveals that only the loss-invisible overlaps enable robust decoding of the training history, while the accuracy of the loss-visible set drops to chance level (Fig.~\ref{fig:4}f). This suggests that loss-invisible overlaps can serve as memory variables. 

\section{Related work}
\paragraph{Low-rank RNNs}
RNNs are widely used in theoretical neuroscience and machine learning~\cite{barak2017recurrent,maheswaranathan2019reverse,graves2013speech,pascanu2013difficulty,orvieto2023resurrecting}. A particularly tractable class, especially in theoretical neuroscience, consists of RNNs with low-rank connectivity, which simplifies the dynamics and enables analysis via a small set of macroscopic overlap variables~\cite{mastrogiuseppe2018linking}. Historically, research on low-rank RNNs has followed two main directions. The first constructs networks by hand, designing connectivity to implement specific computations and analyzing the resulting dynamics~\cite{mastrogiuseppe2018linking,schuessler2020dynamics,beiran2021shaping,hopfield1982neural,marschall2025theory,clark2025connectivity}. The second studies how such structure emerges through learning, either by training low-rank networks directly or by showing that unconstrained networks develop low-rank solutions~\cite{schuessler2020interplay,dubreuil2022role,valente2022extracting,pals2024trained}.

\paragraph{Learning dynamics of RNNs} More recently, a third line of work has begun to analyze RNN learning dynamics analytically~\cite{schuessler2020interplay,bordelon2025dynamically,proca2025learning,ger2026learning}. Two particularly relevant studies are~\cite{bordelon2025dynamically,proca2025learning}, which analyze unconstrained \emph{linear} RNNs in both the lazy and rich regimes. In~\cite{bordelon2025dynamically}, learning equations are derived for simplified tasks, building on earlier results~\cite{schuessler2020interplay}. In~\cite{proca2025learning}, ideas from feedforward networks~\cite{saxe2013exact,saxe2019mathematical} are extended to RNNs, strongly relying on task decomposition into singular modes. These approaches obtain tractability through assumptions such as timescale separation, frozen parameters, or balanced, aligned, or tied initializations. Our framework instead leverages the low-rank architecture rather than a task-specific decomposition, allowing all parameters to evolve simultaneously under gradient descent. Within this setting, we further show that initialization structure (including the relative magnitudes across layers, beyond their overall scale~\cite{azulay2021implicit}), particularly within the loss-invisible set, can substantially alter learning trajectories. Finally, we extend the analysis to \emph{nonlinear} RNNs, providing, to our knowledge, the first analytical treatment of their full deterministic gradient-descent trajectories. We note that two recent works, which appeared during the preparation and review of our manuscript~\cite{bauer2026unified,clark2026structure}, provide complementary insights into learning in nonlinear RNNs through Bayesian mean-field descriptions. These works characterize the equilibrium distribution of trained weights under Langevin dynamics rather than the full learning trajectory.

\paragraph{Parameter and function space duality}
Our work analyzes learning dynamics in a low-dimensional overlap space rather than the high-dimensional parameter space, focusing on the function implemented by the network. This perspective is related to the Neural Tangent Kernel (NTK) framework~\cite{jacot2018neural,chizat2019lazy}, where learning in infinitely wide networks (mostly feedforward) is described at the level of functions rather than parameters. However, while NTK analyses typically focus on the output layer, our approach tracks additional quantities beyond it. Our framework is also related to natural-gradient methods~\cite{amari1998natural,pascanu2013revisiting}. In particular, the matrix $\bm G$ maps parameter-space gradient descent to induced dynamics in overlap space, endowing these coordinates with a non-Euclidean geometry. Conceptually, this plays a similar role to a metric or preconditioner in natural-gradient methods, but here it arises directly from the low-rank parameterization instead of defining a distance metric.

\paragraph{Invariants, symmetries, and drift}
Finally, our analysis connects to work on invariants and symmetries in neural network learning dynamics~\cite{du2018algorithmic,kunin2020neural,tanaka2021noether}. We interpret loss-invisible components of connectivity as potential memory variables. In linear networks under gradient flow, conserved quantities constrain these directions and prevent them from storing training history. Breaking these symmetries—via modified learning rules or nonlinearities—removes this constraint, enabling loss-invisible directions to encode past training. We further show that label noise induces a slow drift along these directions, consistent with results that noisy optimization (e.g., SGD) biases solutions toward balanced, minimal-norm, or flatter regions of the solution manifold~\cite{blanc2020implicit,ratzon2024representational,li2021happens}.

\section{Discussion}
Learning is a fundamental property of both biological and artificial neural networks, yet linking structural changes during learning (e.g., synaptic plasticity or weight updates) to changes in network function remains challenging. Here, we leverage the low-rank RNN framework to derive a low-dimensional description of learning under gradient descent. Our central insight is that learning can be captured through a small set of overlaps that separate connectivity into \textit{loss-visible} directions, which determine network function, and \textit{loss-invisible} directions, which shape how learning unfolds. We demonstrate this perspective in both linear and nonlinear RNNs.

\paragraph{Biological implications} This perspective yields several testable predictions for biological learning experiments. First, we introduce the concept of \emph{perturbation-by-learning}, whereby observing how a system learns can reveal structural differences that remain hidden in static recordings of behavior. In this sense, learning trajectories act as probes of circuit structure. Second, we interpret loss-invisible directions as candidate memory variables encoding a circuit’s training history (and potentially also future learning capacity). This suggests that uncovering learning history requires focusing not on synapses that determine current behavior, but on those that are functionally silent yet central to learning. More broadly, this perspective parallels findings that residual neural activity, though behaviorally silent, may carry valuable information~\cite{galgali2023residual,pereira2025neural}, and extends this principle from neural activity to learning dynamics. Finally, inferring the matrix $\bm G$ could guide the design of efficient shaping protocols tailored to the microscopic parameters whose plasticity mediates learning.

\phantomsection
\label{sec:limitations}
\paragraph{Limitations} We note several limitations of our work. First, for analytical tractability, we focus on rank-1 connectivity. While extensions to higher-rank networks are possible (App.~\ref{app:D}), the number of overlaps grows quadratically with rank, leading to increasingly complex dynamics. Second, gradients with respect to the overlaps are not always analytically tractable; although closed-form expressions exist for some tasks (e.g., the filter task; App.~\ref{app:A43}), they may be unavailable in more complex ones. Nevertheless, simulating the resulting low-dimensional overlap dynamics is far more efficient than the full $N$-dimensional ones. Third, our nonlinear analysis assumes that the parameter-vector components remain approximately Gaussian during training, an assumption not strictly preserved (App.~\ref{app:B41}). Finally, we consider purely low-rank connectivity, enforced by training the low-rank factors directly rather than the full recurrent matrix $\bm W$. Many RNN studies instead train $\bm W$ directly or add a full-rank random bulk to the low-rank connectivity, whose incorporation remains an important future direction.

In summary, our work extends the low-rank RNN framework to incorporate learning dynamics within the same low-dimensional description. This perspective provides a tractable link between changes in connectivity and the evolution of network function, offering a principled framework for studying degeneracy, memory, and drift in recurrent networks, with implications for both neuroscience and machine learning theory.

\section*{Acknowledgments} 
We thank an anonymous reviewer for the suggestion of the compact derivation in Appendix D. This work was supported by the Israel Science Foundation (grant No. 1442/21 to OB) and Human Frontiers Science Program (HFSP) research grant (RGP0017/2021 to OB). The funders had no role in study design, data collection and analysis, decision to publish, or preparation of the manuscript.
\section*{Code Availability}
All code was implemented in Python using PyTorch~\cite{paszke2019pytorch} and is available on GitHub:  \url{https://github.com/yoavger/learning_reveals_invisible_structure_lr_rnns}

\bibliographystyle{unsrt}  


\appendix
\renewcommand{\thesection}{\Alph{section}}
\counterwithin{equation}{section}
\renewcommand{\theequation}{\thesection.\arabic{equation}}

\newpage
\section*{Appendix}
The appendix is organized as follows:
\begin{itemize}
\item \textbf{Section~\ref{app:A}} – Full derivation of the linear rank-1 RNN: reduced activity dynamics, overlap learning dynamics, filter task and training details, a detailed comparison of parameter-space and overlap-space learning, gradient-flow invariants, and experiments using alternative learning rules.

\item \textbf{Section~\ref{app:B}} – Full derivation of the nonlinear rank-1 RNN: mean-field reduced activity dynamics, overlap learning dynamics, flip-flop task and training details, experiments testing the limits of the Gaussian assumptions, and full details of the history-dependent memory protocol (A/B $ \rightarrow$ C).
\item \textbf{Section~\ref{app:C}} – Derivation of the augmented $10\times10$ Gram matrix $\bar{\bm G}(\bm\theta)$, and detailed comparison between linear and nonlinear models.
\item \textbf{Section~\ref{app:D}} – Extension to the linear rank-2 RNN, including how overlaps scale with rank.
\end{itemize}

\newpage
\section{Linear rank-1 RNN}
\label{app:A}
We provide here a complete derivation of the linear rank-1 RNN, including both the within-episode dynamics and the across-episode learning dynamics.
\subsection{Within-episode dynamics} 
\label{app:A1}
With rank-1 connectivity $\bm W=\frac{1}{N}\bm u\bm v^\top$ and linear activation $\phi=\mathrm{id}$, the state dynamics becomes 
\begin{equation}
    \dot{\bm h}(t) = -\bm h(t) + \frac{1}{N}\bm u\bm v^\top \bm h(t) + \bm m\,x(t)
    \label{Eq.A1}
\end{equation}
Assuming $\bm h(0)=\bm 0$, the right-hand side of Eq.~\eqref{Eq.A1}
always lies in $\mathrm{span}\{\bm m,\bm u\}$, since
\[\bm u\bm v^\top \bm h(t)\in \mathrm{span}\{\bm u\} \quad \text{and}\quad \bm m\,x(t)\in \mathrm{span}\{\bm m\} \quad  \] 
Therefore $\bm h(t)\in \mathrm{span}\{\bm m,\bm u\}$ for all $t$, and we may write
\begin{equation}
    \bm h(t)=\kappa_m(t)\,\bm m + \kappa_u(t)\,\bm u,
    \qquad
    \bm\kappa(t)=
    \begin{bmatrix}
        \kappa_m(t)\\[2pt]
        \kappa_u(t)
    \end{bmatrix}\in\mathbb{R}^2
\end{equation}
Projecting Eq.~\eqref{Eq.A1} onto this subspace yields the effective dynamics
\begin{equation}
    \dot{\bm\kappa}(t)
    =-\bm\kappa(t)+
    \begin{bmatrix}
        0 & 0\\
        \tfrac{1}{N}\bm v^\top \bm m & \tfrac{1}{N}\bm v^\top \bm u
    \end{bmatrix}
    \bm\kappa(t) +
    \begin{bmatrix}
        1\\[2pt]0
    \end{bmatrix}x(t)
    \label{Eq.A3}
\end{equation}
and readout 
\begin{equation}
    \hat y(t)=\frac{1}{N}\bm z^\top \bm h(t)
    =\frac{1}{N}
    \begin{bmatrix}
        \bm z^\top \bm m & \bm z^\top \bm u
    \end{bmatrix}
    \bm\kappa(t)
    \label{Eq.A4}
\end{equation}

\subsection{Across-episode learning dynamics}
\label{app:A2}
The episode loss depends on the parameters only through the \emph{loss-visible} overlaps
\begin{equation}
\mathcal L =
\mathcal L\bigl(
\sigma_{zm}, \sigma_{zu}, \sigma_{vm}, \sigma_{vu}
\bigr)
\end{equation}

By the chain rule, the gradients with respect to the parameter vectors can be written as
\begin{align}
\frac{\partial \mathcal{L}}{\partial \bm{m}}
&=
\frac{\partial \mathcal{L}}{\partial \sigma_{zm}}
\frac{\partial \sigma_{zm}}{\partial \bm{m}}
+
\frac{\partial \mathcal{L}}{\partial \sigma_{vm}}
\frac{\partial \sigma_{vm}}{\partial \bm{m}}
=
\frac{1}{N}
\left(
\nabla_{zm}\bm{z}
+
\nabla_{vm}\bm{v}
\right)
\\[4pt]
\frac{\partial \mathcal{L}}{\partial \bm{u}}
&=
\frac{\partial \mathcal{L}}{\partial \sigma_{zu}}
\frac{\partial \sigma_{zu}}{\partial \bm{u}}
+
\frac{\partial \mathcal{L}}{\partial \sigma_{vu}}
\frac{\partial \sigma_{vu}}{\partial \bm{u}}
=
\frac{1}{N}
\left(
\nabla_{zu}\bm{z}
+
\nabla_{vu}\bm{v}
\right)
\\[4pt]
\frac{\partial \mathcal{L}}{\partial \bm{v}}
&=
\frac{\partial \mathcal{L}}{\partial \sigma_{vm}}
\frac{\partial \sigma_{vm}}{\partial \bm{v}}
+
\frac{\partial \mathcal{L}}{\partial \sigma_{vu}}
\frac{\partial \sigma_{vu}}{\partial \bm{v}}
=
\frac{1}{N}
\left(
\nabla_{vm}\bm{m}
+
\nabla_{vu}\bm{u}
\right)
\\[4pt]
\frac{\partial \mathcal{L}}{\partial \bm{z}}
&=
\frac{\partial \mathcal{L}}{\partial \sigma_{zm}}
\frac{\partial \sigma_{zm}}{\partial \bm{z}}
+
\frac{\partial \mathcal{L}}{\partial \sigma_{zu}}
\frac{\partial \sigma_{zu}}{\partial \bm{z}}
=
\frac{1}{N}
\left(
\nabla_{zm}\bm{m}
+
\nabla_{zu}\bm{u}
\right)
\end{align}

Under gradient flow, $\dot{\bm\theta}=-\nabla_{\bm\theta}\mathcal L$, the parameter dynamics become
\begin{equation}
\begin{aligned}
\dot{\bm{m}} &= -{N}^{-1}(\nabla_{zm}\,\bm{z} + \nabla_{vm}\,\bm{v})\,\,
\qquad
\dot{\bm{u}} = -{N}^{-1}(\nabla_{zu}\,\bm{z} + \nabla_{vu}\,\bm{v})\\[4pt]
\dot{\bm{v}} &= -{N}^{-1}(\nabla_{vm}\,\bm{m} + \nabla_{vu}\,\bm{u})
\qquad
\dot{\bm{z}} = -{N}^{-1}(\nabla_{zm}\,\bm{m} + \nabla_{zu}\,\bm{u})
\end{aligned}
\label{Eq.A10}
\end{equation}

By the product rule, for any two vectors $\bm{v}$ and $\bm{u}$, the derivative of their inner product is
\begin{equation}
\frac{d}{d\tau}(\bm{v}^\top \bm{u}) = \dot{\bm{v}}^\top \bm{u} + \bm{v}^\top \dot{\bm{u}}
\end{equation}
Applying this to the loss-visible overlaps together with Eq.~\eqref{Eq.A10} yields the induced dynamics
\begin{equation}
\begin{aligned}
\dot{\sigma}_{zm}
&= -(\|\bm{m}\|^2+\|\bm{z}\|^2)\nabla_{zm}
   -\sigma_{mu}\nabla_{zu}
   -\sigma_{zv}\nabla_{vm}\\[4pt]
\dot{\sigma}_{zu}
&= -\sigma_{mu}\nabla_{zm}
   -(\|\bm{u}\|^2+\|\bm{z}\|^2)\nabla_{zu}
   -\sigma_{zv}\nabla_{vu}\\[4pt]
\dot{\sigma}_{vm}
&= -(\|\bm{m}\|^2+\|\bm{v}\|^2)\nabla_{vm}
   -\sigma_{mu}\nabla_{vu}
   -\sigma_{zv}\nabla_{zm}\\[4pt]
\dot{\sigma}_{vu}
&= -\sigma_{mu}\nabla_{vm}
   -(\|\bm{u}\|^2+\|\bm{v}\|^2)\nabla_{vu}
   -\sigma_{zv}\nabla_{zu}
\end{aligned}
\end{equation}
which revealed six additional \emph{loss-invisible} overlaps that are needed to close the learning dynamics 
\begin{equation}
\sigma_{mu},\quad
\sigma_{zv},\quad
\|\bm m\|^2,\quad
\ \|\bm u\|^2,\quad
\ \|\bm v\|^2,\quad
\ \|\bm z\|^2
\end{equation}
Similarly, their evolution under gradient flow is
\begin{equation}
\begin{aligned}
\dot{\sigma}_{mu}
&= -\sigma_{zu}\nabla_{zm}
    -\sigma_{zm}\nabla_{zu}
    -\sigma_{vu}\nabla_{vm}
    -\sigma_{vm}\nabla_{vu}\\[4pt]
\dot{\sigma}_{zv}
&= -\sigma_{vm}\nabla_{zm}
   -\sigma_{vu}\nabla_{zu}
   -\sigma_{zm}\nabla_{vm}
   -\sigma_{zu}\nabla_{vu}\\[4pt]
\dot{\|\bm{m}\|^2}
&= -2(\sigma_{zm}\nabla_{zm}+\sigma_{vm}\nabla_{vm})\\[4pt]
\dot{\|\bm{u}\|^2}
&= -2(\sigma_{zu}\nabla_{zu}+\sigma_{vu}\nabla_{vu})\\[4pt]
\dot{\|\bm{v}\|^2}
&= -2(\sigma_{vm}\nabla_{vm}+\sigma_{vu}\nabla_{vu})\\[4pt]
\dot{\|\bm{z}\|^2}
&= -2(\sigma_{zm}\nabla_{zm}+\sigma_{zu}\nabla_{zu})
\end{aligned}
\label{Eq.A14}
\end{equation}
These equations form a closed $10$-dimensional ODE system in scalar variables, which fully describes the learning trajectory in overlap space.

\subsection{Learning dynamics in matrix form}
\label{app:A3}
The learning dynamics can be written compactly in matrix form by defining
\begin{equation}
\bm\theta=
\begin{bmatrix}
\bm m\\[2pt]\bm u\\[2pt]\bm v\\[2pt]\bm z
\end{bmatrix}\in\mathbb{R}^{4N},
\qquad
\bm\sigma=
\begin{bmatrix}
\sigma_{zm}\\[2pt]\sigma_{zu}\\[2pt]\sigma_{vm}\\[2pt]\sigma_{vu}
\end{bmatrix}\in\mathbb{R}^4,
\qquad
\nabla_{\bm{\sigma}}\mathcal{L}=\frac{\partial\mathcal L}{\partial\bm\sigma}
=\begin{bmatrix}
\nabla_{zm}\\[2pt]
\nabla_{zu}\\[2pt]
\nabla_{vm}\\[2pt]
\nabla_{vu}
\end{bmatrix} \in\mathbb{R}^4
\end{equation}
Let $\bm D$ denote the Jacobian of the overlap map with respect to the parameters
\begin{equation}
\bm D(\bm\theta)=
\frac{\partial\bm\sigma}{\partial\bm\theta}
=
\frac{1}{N}\begin{bmatrix}
\bm z^\top & 0 & 0 & \bm m^\top\\
0 & \bm z^\top & 0 & \bm u^\top\\
\bm v^\top & 0 & \bm m^\top & 0\\
0 & \bm v^\top & \bm u^\top & 0
\end{bmatrix}\in\mathbb{R}^{4\times4N}
\end{equation}
Under gradient flow, $\dot{\bm\theta}=-\bm D(\bm\theta)^\top \nabla_{\bm{\sigma}}\mathcal{L}$, and the induced overlap dynamics follow as
\begin{equation}
\dot{\bm\sigma}
=
\bm D(\bm\theta)\,\dot{\bm\theta}
=
-\bm D(\bm\theta)\bm D(\bm\theta)^\top\,\nabla_{\bm{\sigma}}\mathcal{L}
=
-\bm G(\bm\theta)\,\nabla_{\bm{\sigma}}\mathcal{L},
\qquad
\bm G(\bm\theta)= \bm D(\bm\theta)\bm D(\bm\theta)^\top
\end{equation}
The matrix $\bm G(\bm\theta)$ is given explicitly by
\begin{equation}
\bm{G}(\bm\theta)
= \frac{1}{N}
\begin{bmatrix}
\|\bm m\|^2+\|\bm z\|^2 & \sigma_{mu} & \sigma_{zv} & 0\\
\sigma_{mu} & \|\bm u\|^2+\|\bm z\|^2 & 0 & \sigma_{zv}\\
\sigma_{zv} & 0 & \|\bm m\|^2+\|\bm v\|^2 & \sigma_{mu}\\
0 & \sigma_{zv} & \sigma_{mu} & \|\bm u\|^2+\|\bm v\|^2
\end{bmatrix}\in\mathbb{R}^{4\times4}
\end{equation}
The dynamics of the loss-invisible overlaps can be written in an analogous matrix form. Define
\begin{equation}
\tilde{\bm\sigma}
=
\begin{pmatrix}
\sigma_{mu},\,
\sigma_{zv},\,
\|\bm m\|^2,\,
\|\bm u\|^2,\,
\|\bm v\|^2,\,
\|\bm z\|^2
\end{pmatrix}^{\!\top},
\quad
\tilde{\bm D (\theta)}= \frac{\partial\tilde{\bm\sigma}}{\partial\bm\theta}
\end{equation}
Using $\dot{\bm\theta}=-\bm D^\top(\bm\theta)\nabla_{\bm{\sigma}}\mathcal{L}$, the induced dynamics of $\tilde{\bm\sigma}$ follow as
\begin{equation}
\dot{\tilde{\bm\sigma}}
=
-\tilde{\bm D}(\bm\theta)\bm D(\bm\theta)^\top\,\nabla_{\bm{\sigma}}\mathcal{L}
=
-\tilde{\bm G}(\bm\theta)\,\nabla_{\bm{\sigma}}\mathcal{L},
\qquad
\tilde{\bm G}(\bm\theta)= \tilde{\bm D}(\bm\theta)\bm D^\top(\bm\theta)
\label{app:eq:sigma_tilde_metric_form}
\end{equation}
With $\tilde{\bm G(\bm\theta)}$ given explicitly as
\begin{equation}
\tilde{\bm G}(\bm\theta)
= \frac{1}{N}
\begin{bmatrix}
\sigma_{zu} & \sigma_{zm} & \sigma_{vu} & \sigma_{vm}\\
\sigma_{vm} & \sigma_{vu} & \sigma_{zm} & \sigma_{zu}\\
2\sigma_{zm} & 0 & 2\sigma_{vm} & 0\\
0 & 2\sigma_{zu} & 0 & 2\sigma_{vu}\\
0 & 0 & 2\sigma_{vm} & 2\sigma_{vu}\\
2\sigma_{zm} & 2\sigma_{zu} & 0 & 0
\end{bmatrix}\in\mathbb{R}^{6\times4}
\end{equation}

That is, the learning dynamics reduce to a closed 10D coupled ODE system
\begin{equation}
\dot{\bm\sigma} = -\bm G(\bm\theta)\,\nabla_{\bm{\sigma}}\mathcal{L},
\qquad
\dot{\tilde{\bm\sigma}}= -\tilde{\bm G}(\bm\theta)\,\nabla_{\bm{\sigma}}\mathcal{L}
\end{equation}
which translates parameter-space learning dynamics into overlap-space learning. For a complete treatment, we derive the augmented $10\times10$ Gram matrix $\bar{\bm{G}}$ in App.~\ref{app:C}. 

\subsection{Filter task}
\label{app:A4}
\begin{figure}[!h]
    \centering
\includegraphics[width=\linewidth]{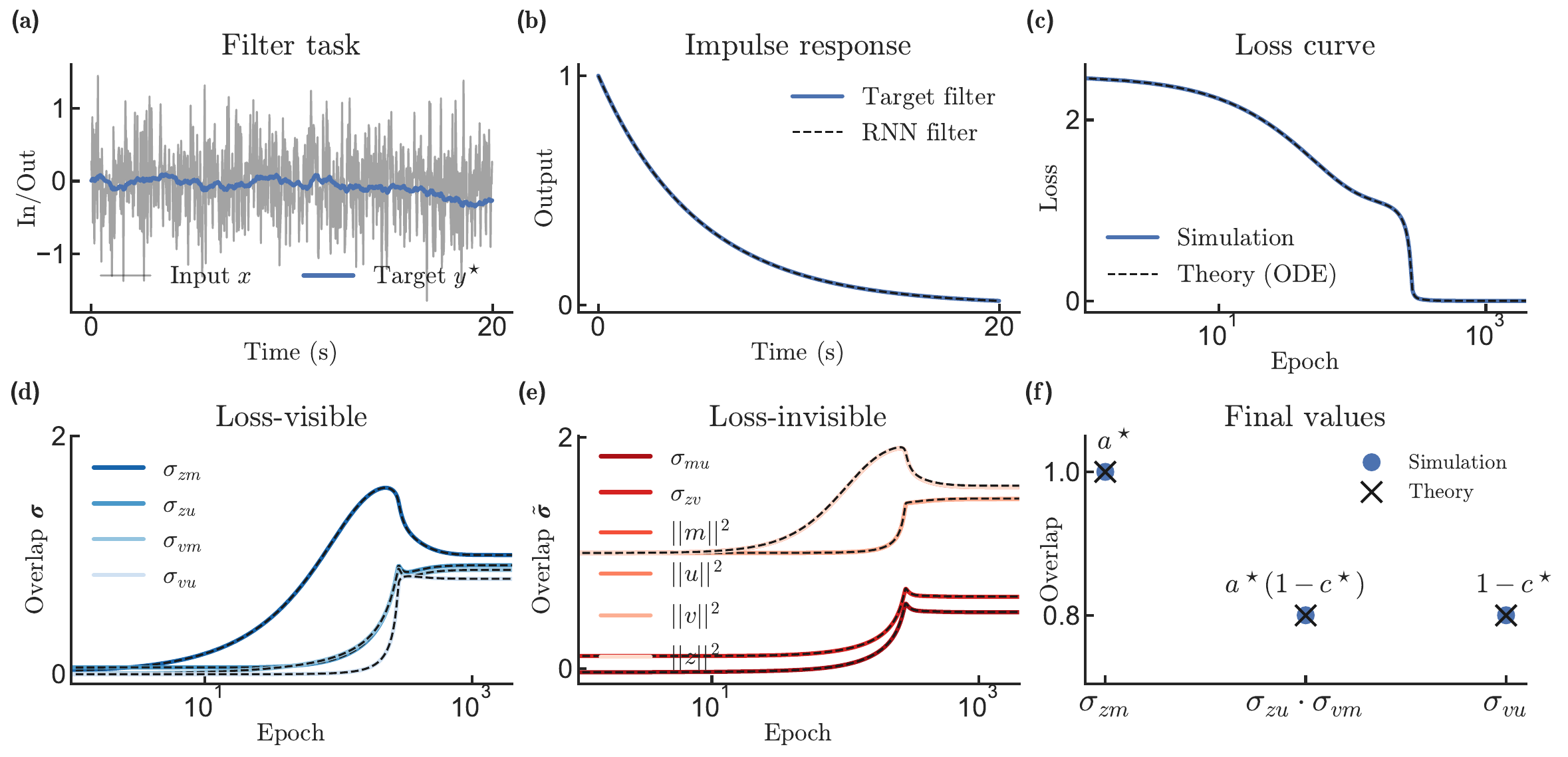}
\caption{\textbf{(a)} Filter task with white-noise input $x$ (gray) and target output $y^\star$ (blue). \textbf{(b)} Impulse responses of the target filter (solid blue), with gain $a^\star=1$ and decay rate $c^\star=0.2$, and final learned RNN function (dashed black).
\textbf{(c)} Training loss of the RNN: numerical simulation of the high-dimensional network (solid blue) and the corresponding ODE theory (dashed black), showing excellent agreement. \textbf{(d,e)} Dynamics of loss-visible and loss-invisible overlaps, comparing numerical simulations (solid) with ODE predictions (dashed), again in close agreement. \textbf{(f)} Final overlap values from simulation (blue circles) compared with theoretical predictions (black crosses).}
\label{fig:5}
\end{figure}

As a concrete example for the rank-1 linear RNN, we consider the simple task of reproducing a first-order exponential filter driven by white-noise input~\cite{bordelon2025dynamically} (Fig.~\ref{fig:5}). The transfer function defines the target response
\begin{equation}
y^\star(t)=a^\star\, e^{-c^\star t} * x(t)   
\qquad\Longleftrightarrow\qquad
H^\star(s)=\frac{a^\star}{s+c^\star} 
\end{equation}
where $x(t)$ denotes a white-noise input signal, $a^\star$ is the filter gain, $c^\star>0$ is the decay rate, and $y^\star(t)$ is the target. 

\subsubsection{Training details and RNN initialization} 
\label{app:A41}
\begin{figure}[!h]
    \centering
    \includegraphics[width=0.9\linewidth]{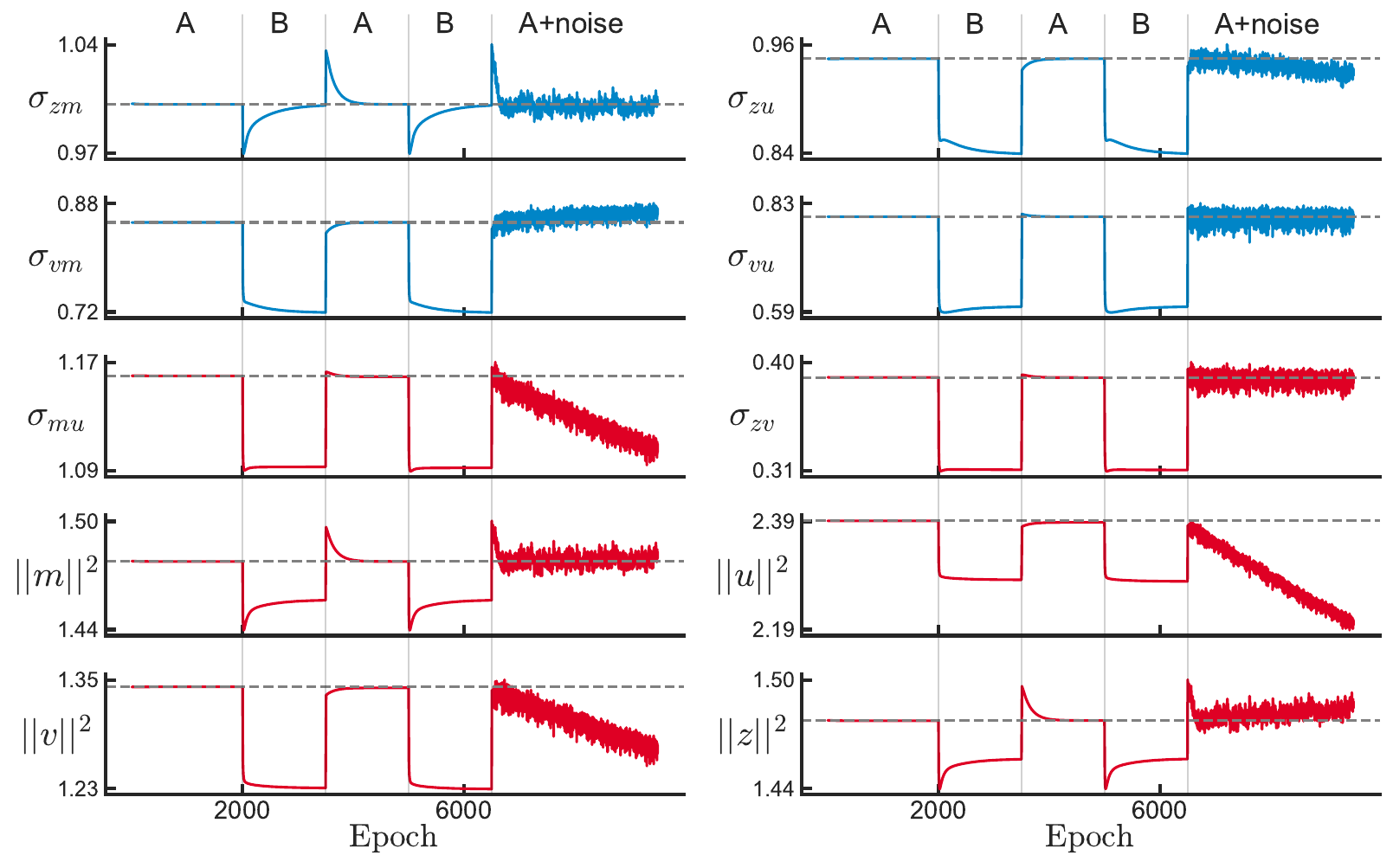}
\caption{Trajectories of all ten overlaps in the A-B-A protocol, supplementing Fig.~\ref{fig:3} of the main text. Blue traces represent loss-visible overlaps, and red traces depict loss-invisible ones. During the initial ABAB phases, all overlaps return to their previous values, demonstrating a lack of memory. However, when noise is added to the training process, we observe a distinct drift, which is most pronounced in the loss-invisible overlaps. However, also note that the loss-visible overlaps can exhibit drift under noisy conditions ($\sigma_{vm} $ and $\sigma_{zu}$); this is understood through the degenerate task condition, where the filter task only constrains the product $\sigma_{vm}\sigma_{zu}$, allowing noise to drive the system toward a balanced solution where these individual overlaps equalize (see App.~\ref{app:A42}).}
\label{fig:7}
\end{figure}

Numerical simulations were performed by training a continuous-time rank-1 RNN discretized via the Euler method ($\Delta t=0.025$) and size $N=500$. Every element of the trainable vectors $\bm{\theta}=\{\bm m,\bm u,\bm v,\bm z\}\subset\mathbb{R}^N$ is initialized i.i.d. from a standard normal distribution $\mathcal N(0,1)$. Training is performed by gradient descent with learning rate $\eta=5\times10^{-3}$ to match the filter target with gain $a^\star=1$ and decay rate $c^\star=0.2$, using the impulse (delta) response as input, and for an episode of length $T=20\,\mathrm{s}$. The loss is the time-integrated squared error between the network output and the target. For the A–B–A training protocol, task A filter parameters are as described above, while task B uses the same gain $a^\star=1$ but a different decay rate $c^\star=0.4$. In the same protocol, during the noisy training phase, Gaussian noise $\mathcal{N}(0, 0.01)$ is added to the target labels (Fig.~\ref{fig:7}).

\subsubsection{Network impulse response and exact solution}
\label{app:A42}
For the effective 2D RNN described in Eqs.~\eqref{Eq.A3},~\eqref{Eq.A4}, the impulse response can be obtained in closed form. Define
\begin{equation}
A=\begin{bmatrix}
-1 & 0\\
\sigma_{vm} & -1+\sigma_{vu}
\end{bmatrix},
\qquad
B=\begin{bmatrix}1\\0\end{bmatrix},
\qquad
C=\begin{bmatrix}\sigma_{zm} & \sigma_{zu}\end{bmatrix}
\end{equation}
The Laplace transform of the transfer function from the input $x$ to the output $\hat y$ reads
\begin{equation}
H(s)
= C (sI-A)^{-1} B
= \frac{s\,\sigma_{zm}+\sigma_{vm}\sigma_{zu}-\sigma_{vu}\sigma_{zm}+\sigma_{zm}}
       {(s+1)(s+1-\sigma_{vu})}
\end{equation}

and the corresponding impulse response (kernel) is a sum of two exponentials
\begin{equation}
\Big(\sigma_{zm}-\frac{\sigma_{zu}\sigma_{vm}}{\sigma_{vu}}\Big)e^{-t}
\;+\;
\frac{\sigma_{zu}\sigma_{vm}}{\sigma_{vu}}\,e^{-(1-\sigma_{vu})t}
\end{equation}
To match the target one–pole filter $a^\star e^{-c^\star t}$ exactly, the unwanted $e^{-t}$ mode must vanish, and the remaining pole and amplitude must be set to $c^\star$ and $a^\star$, respectively. This yields the following constraints
\begin{equation}
\sigma_{vu}=1-c^\star,
\qquad
\frac{\sigma_{zu}\sigma_{vm}}{\sigma_{vu}}=a^\star,
\qquad
\sigma_{zm}=a^\star
\label{Eq.A27}
\end{equation}
Notably, this expression depends on $\sigma_{zu}$ and $\sigma_{vm}$ only through their product, implying a continuous rescaling symmetry
$ \sigma_{zu}\!\to\!\alpha\,\sigma_{zu},\; \sigma_{vm}\!\to\!\alpha^{-1}\sigma_{vm}$ under which the network function is invariant. This further mean, that any pair $(\sigma_{zu},\sigma_{vm})$ satisfying the product constraint $\sigma_{zu}\sigma_{vm}=a^\star(1-c^\star)$ provides a valid solution. These constraints reveal a 1D degenerate manifold of global minima in the visible overlap space. Within this manifold, we identify the \textbf{balanced solution} as the unique symmetric point where
\begin{equation}
\sigma_{zu} = \sigma_{vm} = \sqrt{a^\star(1-c^\star)}
\end{equation}
This point is relevant to our drift results, where training under noisy conditions tends to converge toward this solution.
\subsubsection{Exact gradient calculation}
\label{app:A43}
For the filter task considered here, the gradients with respect to the loss-visible overlaps (i.e., $\nabla_{\sigma}\mathcal{L}$) can be derived in closed form. 

Let $e(t)=\hat y(t)-y^\star(t)$ denote the output error. Writing the impulse-response coefficients as
\begin{equation}
A = \sigma_{zm}-\frac{\sigma_{zu}\sigma_{vm}}{\sigma_{vu}},
\qquad
B = \frac{\sigma_{zu}\sigma_{vm}}{\sigma_{vu}}
\end{equation}
the output takes the form
\begin{equation}
\hat y(t)=A\,e^{-t}+B\,e^{-(1-\sigma_{vu})t}
\end{equation}
Define the difference
\begin{equation}
\delta(t)= e^{-(1-\sigma_{vu})t}-e^{-t}
\end{equation}
The sensitivities of the output with respect to the loss-visible overlaps are
\begin{equation}
\begin{aligned}
\partial_{\sigma_{zm}}\hat y(t)
&= e^{-t}\\
\partial_{\sigma_{zu}}\hat y(t)
&= \frac{\sigma_{vm}}{\sigma_{vu}}\,\delta(t)\\
\partial_{\sigma_{vm}}\hat y(t)
&= \frac{\sigma_{zu}}{\sigma_{vu}}\,\delta(t)\\
\partial_{\sigma_{vu}}\hat y(t)
&=
B\!\left[
-\frac{1}{\sigma_{vu}}\delta(t)
+t\,e^{-(1-\sigma_{vu})t}
\right]
\end{aligned}
\end{equation}
For the episode loss $\mathcal L= \int_{0}^{T} \bigl[\hat{y}(t) - y^\star(t)\bigr]^2\,dt = \int_0^T e(t)^2\,dt$, the corresponding gradients are
\begin{equation}
\begin{aligned}
\nabla_{zm}
&= \partial_{\sigma_{zm}}\mathcal L
=
2\!\int_0^T e(t)\,e^{-t}\,dt\\
\nabla_{zu}
&= \partial_{\sigma_{zu}}\mathcal L
=
2\!\int_0^T e(t)\,\frac{\sigma_{vm}}{\sigma_{vu}}\delta(t)\,dt\\
\nabla_{vm}
&= \partial_{\sigma_{vm}}\mathcal L
=
2\!\int_0^T e(t)\,\frac{\sigma_{zu}}{\sigma_{vu}}\delta(t)\,dt\\
\nabla_{vu}
&= \partial_{\sigma_{vu}}\mathcal L
=
2\!\int_0^T e(t)\,
B\!\left[
-\frac{1}{\sigma_{vu}}\delta(t)
+t\,e^{-(1-\sigma_{vu})t}
\right]dt
\end{aligned}
\end{equation}

\newpage

\subsection{Differences between parameter-space and overlap-space learning}
\label{app:A5}
\begin{figure}[!h]
    \centering
    \includegraphics[width=\linewidth]{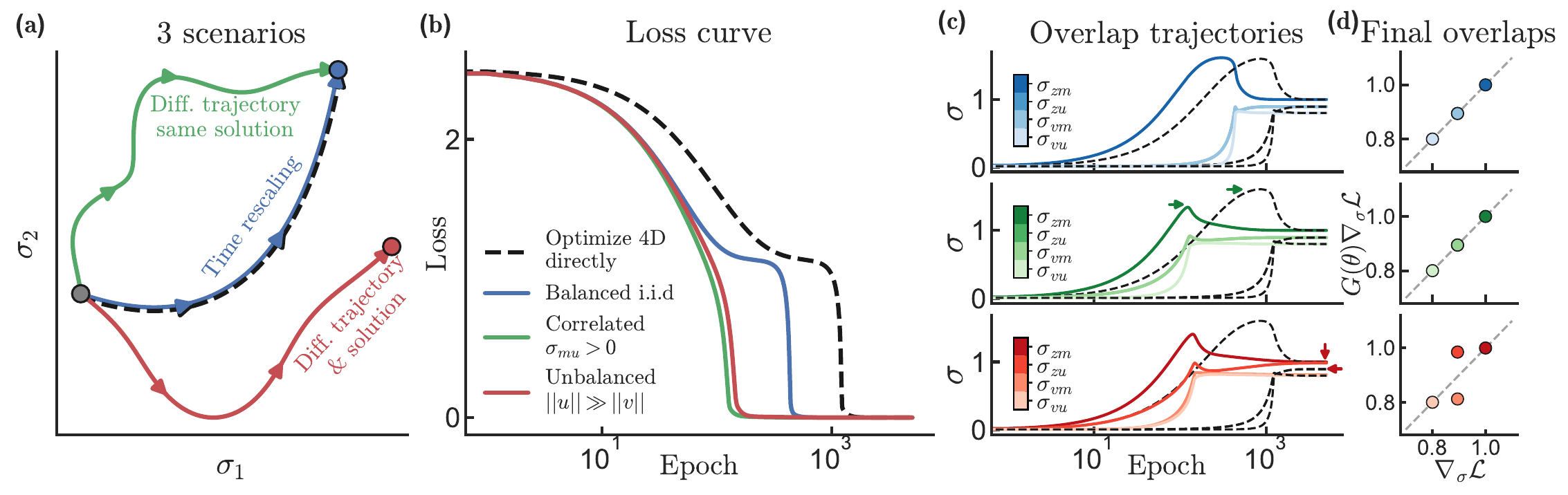}
    \caption{\textbf{(a)} Schematic of three possible regimes. Relative to direct overlap-space optimization (black dashed), the induce parameter-space learning may follow the same trajectory up to a rescaling of time (blue), a different trajectory to the same solution (green), or a different trajectory to a different solution (red). \textbf{(b)} Concrete examples of these regimes. Loss curves for the filter task shown for direct optimization in overlap space (black dashed) and for parameter-space induced dynamics under balanced i.i.d.\ (blue), correlated $\sigma_{mu}>0$ (green), and unbalanced $\|\bm u\|\gg\|\bm v\|$ (red) initializations. \textbf{(c)} Corresponding trajectories of the four loss-visible overlaps for the balanced (top), correlated (middle), and unbalanced (bottom) initializations, shown alongside direct overlap-space optimization (black dashed). Arrows highlight differences between the induced and direct trajectories. \textbf{(d)} Final overlap values under direct optimization (x-axis) and parameter-space induced dynamics (y-axis). Each point denotes one loss-visible overlap, and the dashed diagonal denotes the identity. The balanced and correlated initializations converge to the same solution as direct optimization, whereas the unbalanced initialization selects a different solution, as indicated by the off-diagonal values.}
    \label{fig:6}
\end{figure}

As noted in the main text, parameter-space gradient flow induces in overlap space the following dynamics $\dot{\bm\sigma}=-\bm G(\bm\theta)\nabla_{\bm\sigma}\mathcal L$, whereas direct overlap-space optimization follows $\dot{\bm\sigma}=-\nabla_{\bm\sigma}\mathcal L$. The two dynamics follow the same path up to a rescaling of time whenever $\bm G(\bm\theta)\nabla_{\bm\sigma}\mathcal L=\alpha(t)\nabla_{\bm\sigma}\mathcal L$ for some $\alpha(t)>0$, with $\bm G=\alpha\bm I$ as a sufficient condition. More generally, however, anisotropic diagonal entries or off-diagonal mixing in $\bm G(\theta
)$ can alter the trajectory and the converged solution. To illustrate these possibilities, Fig.~\ref{fig:6} compares parameter-induced and direct overlap-space learning dynamics on the linear filter task (see~\ref{app:A4}) across three initialization regimes. For a balanced i.i.d.\ initialization, the metric is effectively isotropic, and the dynamics differ only by a time rescaling (blue). Introducing initial correlations through $\sigma_{mu}>0$ changes the trajectory but not the final solution (green), whereas an unbalanced initialization with $\|\bm u\|\gg\|\bm v\|$ changes both the trajectory and solution selected on the degenerate manifold (red).

\newpage
\subsection{An invariant under gradient flow}
\label{app:A6}
Define the \(N\times 2\) matrices
\begin{equation}
\bm A = \begin{bmatrix}\bm z & \bm v\end{bmatrix},
\qquad
\bm B = \begin{bmatrix}\bm m & \bm u\end{bmatrix}
\end{equation}
Note that $\bm A$ and $\bm B$ are disjoint, which will be important for the derivation below (and does not hold in the nonlinear case App.~\ref{app:B}). The four loss-visible overlaps assemble into the $2\times 2$ matrix
\begin{equation}
\frac{1}{N}\bm A^\top \bm B
=
\frac{1}{N}\begin{bmatrix}
\bm z^\top\bm m & \bm z^\top\bm u\\[2pt]
\bm v^\top\bm m & \bm v^\top\bm u
\end{bmatrix}
=
\begin{bmatrix}
\sigma_{zm} & \sigma_{zu}\\[2pt]
\sigma_{vm} & \sigma_{vu}
\end{bmatrix}
\end{equation}
Since the within-episode dynamics, readout, and loss depend on the parameters only through $\tfrac{1}{N}\bm A^\top \bm B$, the loss can be written as
\begin{equation}
\mathcal L=\mathcal L(\tfrac{1}{N}\bm A^\top \bm B)
\end{equation}
Defining the $2\times 2$ gradient matrix
\begin{equation}
\bm J = \frac{\partial \mathcal L}{\partial(\bm A^\top \bm B)}
=
\begin{bmatrix}
\nabla_{zm} & \nabla_{zu}\\[2pt]
\nabla_{vm} & \nabla_{vu}
\end{bmatrix}
\end{equation}
gradient flow in parameter space takes the form
\begin{equation}
\dot{\bm A}=-\bm B\,\bm J^\top,
\qquad
\dot{\bm B}=-\bm A\,\bm J
\label{Eq.A38}
\end{equation}
which reproduces Eq.~\eqref{Eq.A10}. Now, consider the matrix difference $\bm A\bm A^\top-\bm B\bm B^\top\in\mathbb R^{N\times N}$
\begin{equation}
\frac{d}{d\tau}(\bm A\bm A^\top)
=
-\bm B\bm J^\top\bm A^\top-\bm A\bm J\bm B^\top
\label{Eq.A35}
\end{equation}
and similarly
\begin{equation}
\frac{d}{d\tau}(\bm B\bm B^\top)
=
-\bm A\bm J\bm B^\top-\bm B\bm J^\top\bm A^\top
\label{Eq.A36}
\end{equation}
Since the RHS of Eqs.~\eqref{Eq.A35},~\eqref{Eq.A36} are identical we have that
\begin{equation}
\frac{d}{d\tau}\big(\bm A\bm A^\top-\bm B\bm B^\top\big)=0
\end{equation}
Therefore the matrix
\begin{equation}
\bm K = \bm A\bm A^\top-\bm B\bm B^\top
= \bm z\bm z^\top+\bm v\bm v^\top-\bm m\bm m^\top-\bm u\bm u^\top
\end{equation}
is conserved under gradient flow and is fixed entirely by the initialization $\big(\bm A(0),\bm B(0)\big)$.

\paragraph{Scalar Invariants}
While $\bm{K} \in \mathbb{R}^{N \times N}$ is a high-dimensional matrix, it is constructed from only four vectors, implying $\mathrm{rank}(\bm{K}) \le 4$. Consequently, its conservation provides exactly four independent scalar constraints, given by the traces of its powers
\[
\mathcal{C}_k = \mathrm{Tr}(\bm{K}^k), \qquad k=1,\dots,4
\]
Using the cyclic property of the trace, these invariants can be expressed directly in terms of the vector norms and pairwise overlaps. The first invariant ($k=1$) is given by the squared norms of all connectivity vectors
\begin{equation}
\mathcal{C}_1 = \mathrm{Tr}(\bm{K})
= \|\bm{z}\|^2 + \|\bm{v}\|^2 - \|\bm{m}\|^2 - \|\bm{u}\|^2 
\end{equation}

The second invariant ($k=2$) couples the norms and all overlaps
\begin{equation}
\mathcal{C}_2 =
\left(\|\bm{z}\|^4 + \|\bm{v}\|^4 + 2\sigma_{zv}^2\right)
+
\left(\|\bm{m}\|^4 + \|\bm{u}\|^4 + 2\sigma_{mu}^2\right)
-
2\left(
\sigma_{zm}^2 + \sigma_{zu}^2 + \sigma_{vm}^2 + \sigma_{vu}^2
\right)
\end{equation}
Similarly, $\mathcal{C}_3$ and $\mathcal{C}_4$ expand into increasingly complex overlap combinations. Collectively, these four scalar invariants constrain the 10-dimensional overlap system to a 6-dimensional manifold. From the gradient flow Eq.~\eqref{Eq.A14}, the symmetry in the updates further implies that the differences $\|\bm{z}\|^2 - \|\bm{m}\|^2$ and $\|\bm{v}\|^2 - \|\bm{u}\|^2$ remain constant whenever $\sigma_{vm}\nabla_{vm} = \sigma_{zu}\nabla_{zu}$, a condition satisfied for the filter task, due to the symmetric dependence of the loss on the product $\sigma_{zu}\sigma_{vm}$ (see Eq.~\eqref{Eq.A27}). Under these combined structural and task-specific constraints, there are insufficient degrees of freedom for the six loss-invisible overlaps to drift independently. Their evolution is thus tied to the initialization $\bm K$, leading to an exact recovery observed in the A–B–A protocol.

\begin{figure}[!h]
    \centering
    \includegraphics[width=0.9\linewidth]{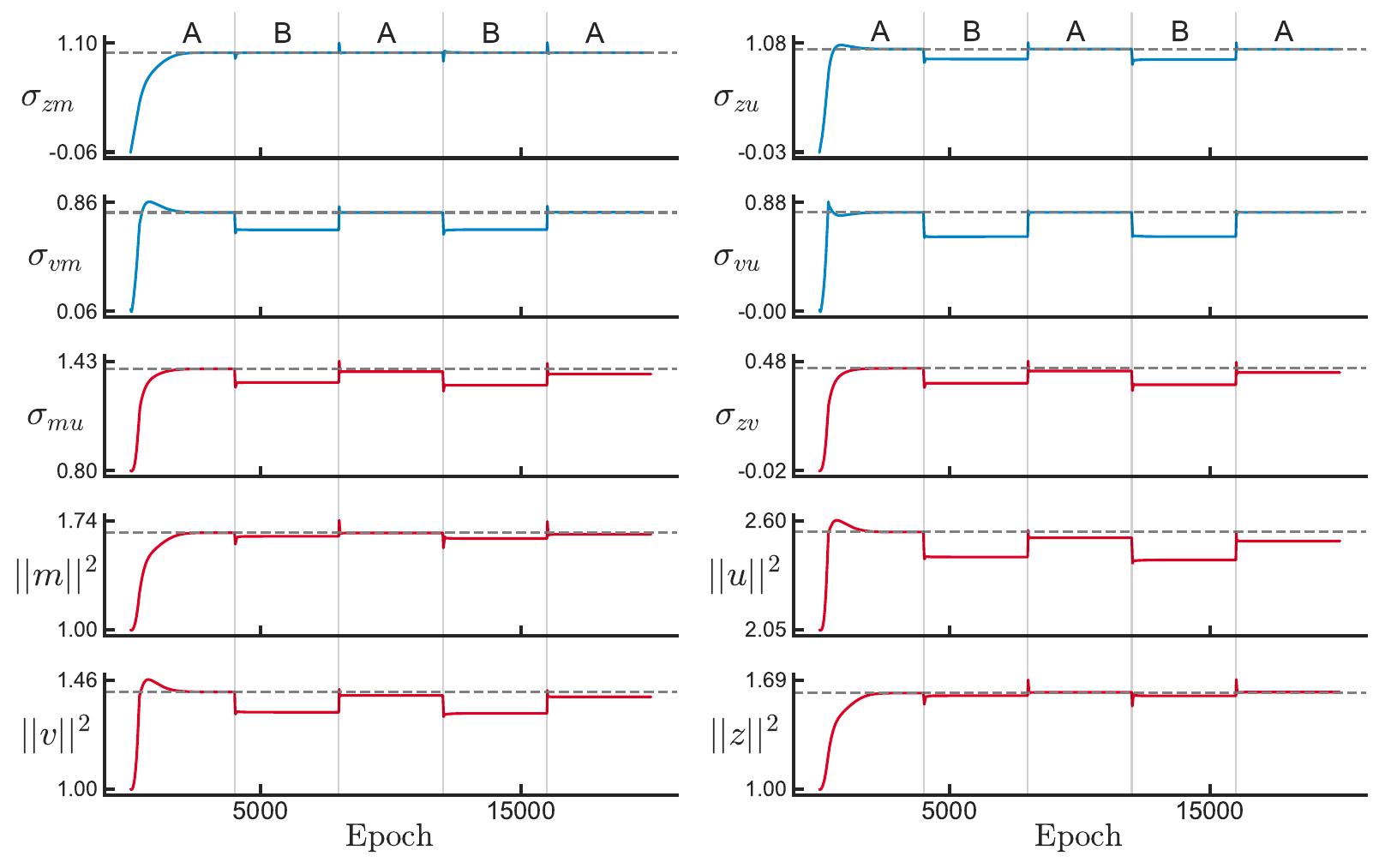}
\caption{Trajectories of all ten overlaps in the A-B-A protocol using the Adam optimizer ($\eta=10^{-3}$). Blue traces represent loss-visible overlaps, and red traces depict loss-invisible ones. Note that, unlike vanilla gradient descent, the adaptive optimizer violates the learning invariant, allowing the loss-invisible overlaps to settle to new values upon retraining, demonstrating history-dependent (red traces do not return to the gray dashed baseline).}
\label{fig:8}
\end{figure}

\section{Nonlinear rank-1 RNN}
\label{app:B}
We provide here a complete derivation of the learning dynamics of the nonlinear rank-1 RNN.
\subsection{Within-episode dynamics}
We consider a rank-1 recurrent network with nonlinear activation
\begin{equation}
\dot{\bm h}(t) = -\bm h(t) + \frac{1}{N}\bm u\bm v^\top \phi(\alpha \bm h(t)) + \bm m\,x(t), \qquad \phi(x)=\mathrm{erf}(x), \qquad \alpha=\sqrt{\pi}/2 
\end{equation} 
where the choice $\alpha=\sqrt{\pi}/2$ ensures that the activation is locally linear with unit slope at the origin. As in the linear case, the recurrent term lies in $\mathrm{span}\{\bm u\}$ and the input term lies in $\mathrm{span}\{\bm m\}$. Assuming $\bm h(0)=\bm 0$, the state remains confined to
\begin{equation}
\bm h(t) = \kappa_m(t)\bm m+\kappa_u(t)\bm u, \qquad \bm\kappa(t)=
\begin{bmatrix} \kappa_m(t)\\[2pt] \kappa_u(t) \end{bmatrix} \in\mathbb R^2
\end{equation}

Assuming that the parameter vectors have zero-mean, the activity also has zero mean, with empirical variance
\begin{equation}
\Delta(t) = \frac{1}{N}\bm h(t)^\top\bm h(t) = \kappa_m^2\|\bm m\|^2 +\kappa_u^2\|\bm u\|^2 +2\kappa_m\kappa_u\sigma_{mu}
\end{equation}
In the large-$N$ limit, neuron averages concentrate to expectations
\begin{equation}
\frac{1}{N}\bm v^\top\phi(\alpha\bm h) = \frac{1}{N}\sum_i v_i\phi(\alpha h_i) \xrightarrow{N\to\infty} \mathbb E[v_i\phi(\alpha h_i)]
\end{equation}
Following the dynamical mean-field theory (DMFT) framework~\cite{sompolinsky1988chaos,mastrogiuseppe2018linking}, we assume that the components of the parameter vectors are jointly Gaussian. Since $h_i(t)=\kappa_m(t)m_i+\kappa_u(t)u_i$, it follows that $h_i(t)\sim\mathcal N(0,\Delta(t))$. Applying Stein's lemma to $(v_i,h_i)$ gives
\begin{equation}
\mathbb E[v_i\phi(\alpha h_i)] = \operatorname{Cov}(v_i,h_i)\,\langle\phi'\rangle
\end{equation}
where, for the error-function nonlinearity, that gain is given in closed-form
\begin{equation}
\langle\phi'\rangle = \mathbb E_{g\sim\mathcal N(0,\Delta(t))} \left[\alpha\phi'(\alpha g)\right] = \left(1+\frac{\pi}{2}\Delta(t)\right)^{-1/2}
\end{equation}
Using $h_i=\kappa_m m_i+\kappa_u u_i$ and the zero-mean assumption
\begin{equation}
\operatorname{Cov}(v_i,h_i) =  \kappa_m\operatorname{Cov}(v_i,m_i) +\kappa_u\operatorname{Cov}(v_i,u_i) = \sigma_{vm}\kappa_m+\sigma_{vu}\kappa_u
\end{equation}
Combining all together yields
\begin{equation}
\frac{1}{N}\bm v^\top\phi(\alpha\bm h) = \left( \sigma_{vm}\kappa_m+\sigma_{vu}\kappa_u \right)\langle\phi'\rangle
\end{equation} 

Substituting into the dynamics, the $N$-dimensional system reduces to
\begin{equation}
\begin{aligned}
\dot\kappa_m(t) &= -\kappa_m(t)+x(t)\\
\dot\kappa_u(t) &= -\kappa_u(t) + \left( \sigma_{vm}\kappa_m(t)+\sigma_{vu}\kappa_u(t) \right)\langle\phi'\rangle
\end{aligned}
\end{equation}
Applying the same argument to $(z_i,h_i)$ gives
\begin{equation}
\hat y(t) = \left( \sigma_{zm}\kappa_m(t)+\sigma_{zu}\kappa_u(t) \right)\langle\phi'\rangle
\end{equation}
Note that, unlike in the linear case, the variance $\Delta(t)$ depends on the overlap $\sigma_{mu}$ as well as the norms $\|\bm m\|^2$ and $\|\bm u\|^2$. As a result, these quantities directly influence the network dynamics and output, and therefore become \emph{loss-visible}.

\newpage
\subsection{Across-episode learning dynamics}
The episode loss depends on the parameters only through the new set of loss-visible overlaps
\begin{equation}
\mathcal L =
\mathcal L\bigl(
\sigma_{zm},\sigma_{zu},\sigma_{vm},\sigma_{vu},
\sigma_{mu},
\| \bm m\|^2,\|\bm u\|^2
\bigr)
\end{equation}

Using $\nabla_{mm}= \partial \mathcal L/\partial \|\bm m\|^2$ and $\nabla_{uu}= \partial \mathcal L/\partial \|\bm u\|^2$, together with $\partial\|\bm m\|^2/\partial\bm m=2\bm m/N$ and $\partial\|\bm u\|^2/\partial\bm u=2\bm u/N$, the chain rule gives
\begin{align}
\frac{\partial \mathcal L}{\partial \bm m}
&= \frac{1}{N}(\nabla_{zm}\,\bm z+\nabla_{vm}\,\bm v+\nabla_{mu}\,\bm u
+2\nabla_{mm}\,\bm m) \\[4pt]
\frac{\partial \mathcal L}{\partial \bm u}
&= \frac{1}{N}(\nabla_{zu}\,\bm z+\nabla_{vu}\,\bm v+\nabla_{mu}\,\bm m
+2\nabla_{uu}\,\bm u) \\[4pt]
\frac{\partial \mathcal L}{\partial \bm z}
&= \frac{1}{N}(\nabla_{zm}\,\bm m+\nabla_{zu}\,\bm u) \\
\frac{\partial \mathcal L}{\partial \bm v}
&= \frac{1}{N}(\nabla_{vm}\,\bm m+\nabla_{vu}\,\bm u)
\nonumber
\end{align}

Under gradient flow $\dot{\bm\theta}=-\nabla_{\bm\theta}\mathcal L$ we obtain
\begin{equation}
\begin{aligned}
\dot{\bm m} &= -\frac{1}{N}(\nabla_{zm}\bm z+\nabla_{vm}\bm v+\nabla_{mu}\bm u+2\nabla_{mm}\bm m) \\[4pt]
\dot{\bm u} &= -\frac{1}{N}(\nabla_{zu}\bm z+\nabla_{vu}\bm v+\nabla_{mu}\bm m+2\nabla_{uu}\bm u) \\[4pt]
\dot{\bm v} &= -\frac{1}{N}(\nabla_{vm}\bm m+\nabla_{vu}\bm u)\\[4pt]
\dot{\bm z} &= -\frac{1}{N}(\nabla_{zm}\bm m+\nabla_{zu}\bm u)
\end{aligned}
\end{equation}

By the product rule of the overlaps, we obtain the dynamics for the loss-visible overlaps
\begin{equation}
\begin{aligned}
\dot\sigma_{zm}
&=
-(\|\bm m\|^2+\|\bm z\|^2)\nabla_{zm}
-\sigma_{mu}\nabla_{zu}
-\sigma_{zv}\nabla_{vm}
-\sigma_{zu}\nabla_{mu}
-2\nabla_{mm}\sigma_{zm}
\\[4pt]
\dot\sigma_{zu}
&=
-\sigma_{mu}\nabla_{zm}
-(\|\bm u\|^2+\|\bm z\|^2)\nabla_{zu}
-\sigma_{zv}\nabla_{vu}
-\sigma_{zm}\nabla_{mu}
-2\nabla_{uu}\sigma_{zu}
\\[4pt]
\dot\sigma_{vm}
&=
-(\|\bm m\|^2+\|\bm v\|^2)\nabla_{vm}
-\sigma_{mu}\nabla_{vu}
-\sigma_{zv}\nabla_{zm}
-\sigma_{vu}\nabla_{mu}
-2\nabla_{mm}\sigma_{vm}
\\[4pt]
\dot\sigma_{vu}
&=
-\sigma_{mu}\nabla_{vm}
-(\|\bm u\|^2+\|\bm v\|^2)\nabla_{vu}
-\sigma_{zv}\nabla_{zu}
-\sigma_{vm}\nabla_{mu}
-2\nabla_{uu}\sigma_{vu}
\\[4pt]
\dot\sigma_{mu}
&=
-\sigma_{zu}\nabla_{zm}
-\sigma_{zm}\nabla_{zu}
-\sigma_{vu}\nabla_{vm}
-\sigma_{vm}\nabla_{vu}
-(\|\bm m\|^2+\|\bm u\|^2)\nabla_{mu}
-2(\nabla_{mm}+\nabla_{uu})\sigma_{mu}
\\[4pt]
\dot{\|\bm m\|^2}
&=
-2\Bigl(
\sigma_{zm}\nabla_{zm}
+\sigma_{vm}\nabla_{vm}
+\sigma_{mu}\nabla_{mu}
+2\nabla_{mm}\|\bm m\|^2
\Bigr)\\[4pt]
\dot{\|\bm u\|^2}
&=
-2\Bigl(
\sigma_{zu}\nabla_{zu}
+\sigma_{vu}\nabla_{vu}
+\sigma_{mu}\nabla_{mu}
+2\nabla_{uu}\|\bm u\|^2
\Bigr)
\end{aligned}
\end{equation}
as well as the loss-invisible dynamics
\begin{equation}
\begin{aligned}
\dot\sigma_{zv}
&=
-\sigma_{vm}\nabla_{zm}
-\sigma_{vu}\nabla_{zu}
-\sigma_{zm}\nabla_{vm}
-\sigma_{zu}\nabla_{vu}\\[4pt]
\dot{\|\bm v\|^2}
&=
-2\Bigl(
\sigma_{vm}\nabla_{vm}
+\sigma_{vu}\nabla_{vu}
\Bigr)\\[4pt]
\dot{\|\bm z\|^2}
&=
-2\Bigl(
\sigma_{zm}\nabla_{zm}
+\sigma_{zu}\nabla_{zu}
\Bigr)
\end{aligned}
\end{equation}

\newpage
\subsection{Learning dynamics in matrix form}
Define 
\begin{equation}
\bm\sigma
=
\begin{bmatrix}
\sigma_{zm}\\
\sigma_{zu}\\
\sigma_{vm}\\
\sigma_{vu}\\
\sigma_{mu}\\
\|\bm m\|^2\\
\|\bm u\|^2
\end{bmatrix},
\qquad
\tilde{\bm\sigma}
=
\begin{bmatrix}
\sigma_{zv}\\
\|\bm v\|^2\\
\|\bm z\|^2
\end{bmatrix},
\qquad
\nabla_{\bm{\sigma}}\mathcal{L}
=
\frac{\partial \mathcal L}{\partial \bm\sigma}
=
\begin{bmatrix}
\nabla_{zm}\\
\nabla_{zu}\\
\nabla_{vm}\\
\nabla_{vu}\\
\nabla_{mu}\\
\nabla_{\|\bm m\|^2}\\
\nabla_{\|\bm u\|^2}
\end{bmatrix}
\label{app:eq:erf_sigma_vec_def}
\end{equation}

The loss-visible Gram matrix $\bm G\in\mathbb R^{7\times 7}$ is
\begin{equation}
\begin{aligned}
\setlength{\arraycolsep}{1pt}
\bm G(\bm\theta)
= \frac{1}{N}
\begin{bmatrix}
\|\bm m\|^2+\|\bm z\|^2 & \sigma_{mu} & \sigma_{zv} & 0 & \sigma_{zu} & 2\sigma_{zm} & 0\\
\sigma_{mu} & \|\bm u\|^2+\|\bm z\|^2 & 0 & \sigma_{zv} & \sigma_{zm} & 0 & 2\sigma_{zu}\\
\sigma_{zv} & 0 & \|\bm m\|^2+\|\bm v\|^2 & \sigma_{mu} & \sigma_{vu} & 2\sigma_{vm} & 0\\
0 & \sigma_{zv} & \sigma_{mu} & \|\bm u\|^2+\|\bm v\|^2 & \sigma_{vm} & 0 & 2\sigma_{vu}\\
\sigma_{zu} & \sigma_{zm} & \sigma_{vu} & \sigma_{vm} & \|\bm m\|^2+\|\bm u\|^2 & 2\sigma_{mu} & 2\sigma_{mu}\\
2\sigma_{zm} & 0 & 2\sigma_{vm} & 0 & 2\sigma_{mu} & 4\|\bm m\|^2 & 0\\
0 & 2\sigma_{zu} & 0 & 2\sigma_{vu} & 2\sigma_{mu} & 0 & 4\|\bm u\|^2
\end{bmatrix}
\end{aligned}
\end{equation}
and the loss-invisible matrix $\tilde{\bm G}\in\mathbb R^{3\times 7}$ is
\begin{equation}
\tilde{\bm G}(\bm \theta) = \frac{1}{N}
\begin{bmatrix}
\sigma_{vm} & \sigma_{vu} & \sigma_{zm} & \sigma_{zu} & 0 & 0 & 0\\
0 & 0 & 2\sigma_{vm} & 2\sigma_{vu} & 0 & 0 & 0\\
2\sigma_{zm} & 2\sigma_{zu} & 0 & 0 & 0 & 0 & 0
\end{bmatrix}
\end{equation}

Using these definitions, the learning dynamics can be written compactly as
\begin{equation}
\dot{\bm\sigma} = -\bm G(\bm\theta)\,\nabla_{\bm{\sigma}}\mathcal{L},
\qquad
\dot{\tilde{\bm\sigma}} = -\tilde{\bm G}(\bm\theta)\,\nabla_{\bm{\sigma}}\mathcal{L}
\end{equation}

\subsection{Flip-flop task}
\label{app:B4}
For our nonlinear presentation, we consider the 1-bit flip-flop task \cite{sussillo2013opening} (Fig.~\ref{fig:10}d). The task requires the network to act as a bistable memory: it must maintain a constant output and only "flip" its state upon receiving a brief, signed input pulse. This behavior requires the creation of stable fixed points separated by a nonlinear boundary. During each trial, the network receives a sequence of short input pulses, each of duration $t_{\mathrm{stim}}$. During a pulse, the input channel is set to $x(t) = s\,x_{\mathrm{amp}}$, where $x_{\mathrm{amp}}=1$, and the sign $s \in \{\pm 1\}$ is chosen at random. Each pulse is followed by a delay period of duration $t_{\mathrm{delay}}$, after which a decision period begins. During this decision period, the loss is activated (i.e., a mask is set to 1), and the target value is defined as ${y}(t) = s\,{y}_{\mathrm{amp}}$, with $y_{\mathrm{amp}}=0.5$. The decision period ends when the next pulse begins. The inter-stimulus delays $t_{\mathrm{isd}}$ are drawn randomly.

\subsubsection{Training details, RNN initialization and Gaussian assumption}
\label{app:B41}
Numerical simulations were performed by training a continuous-time rank-1 RNN discretized Euler method with time step $\Delta t = 0.025$ and network size $N = 1000$. Every element of the trainable vectors $\bm{\theta}=\{\bm m,\bm u,\bm v,\bm z\}\subset\mathbb{R}^N$ is initialized i.i.d. from a standard normal distribution $\mathcal N(0,1)$. Training is performed using gradient descent with learning rate $\eta = 0.05$ over episodes of length $T=20\,\mathrm{s}$ and batch size of 10. We use a masked mean-squared error (MSE) loss that ignores the output during input pulses and short transients, thereby focusing learning on maintaining stable fixed-point outputs. 

\begin{figure}[!h]
\includegraphics[width=\linewidth]{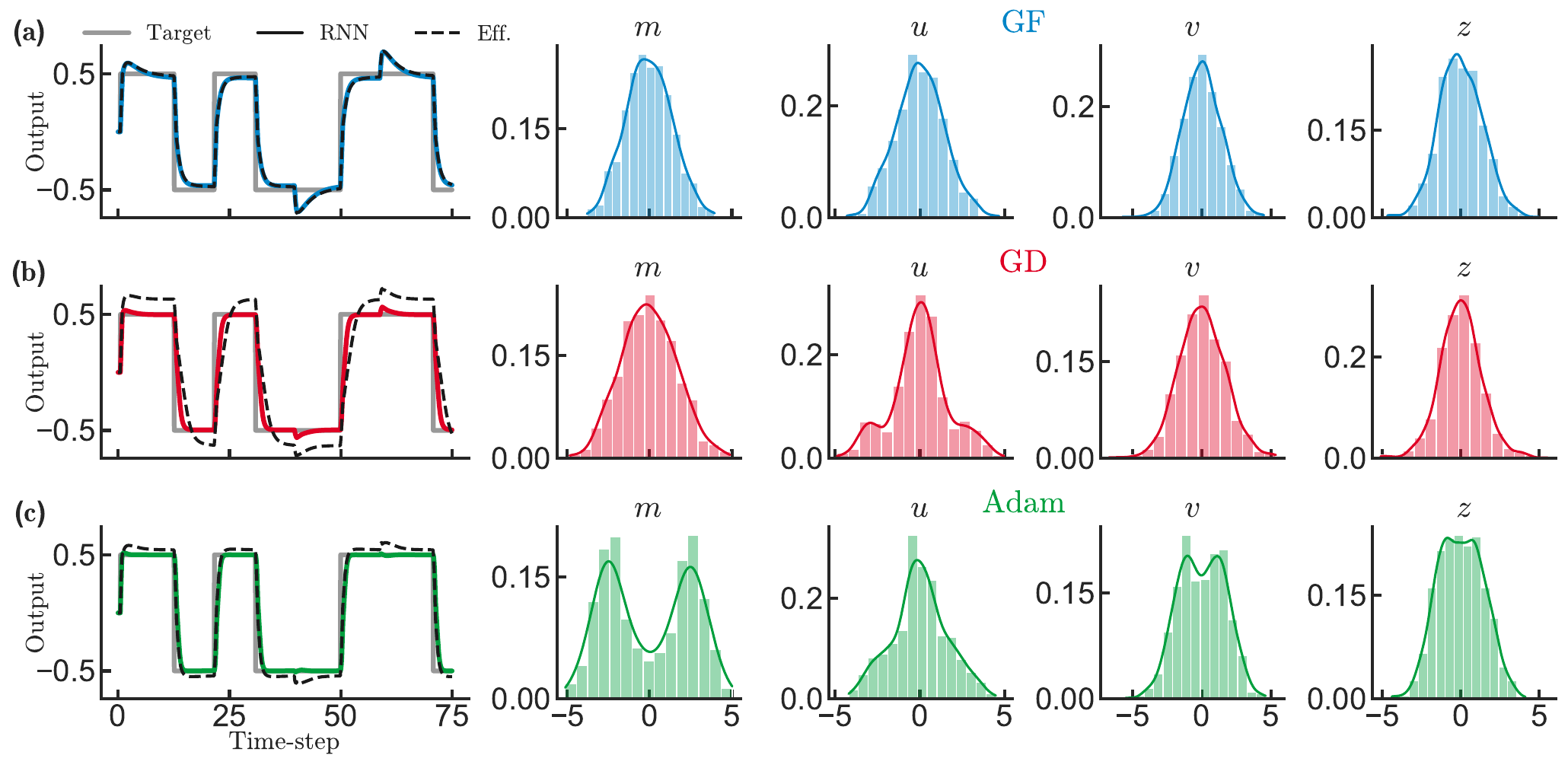}
\caption{Training on flip-flop task with different optimizers. \textbf{(a)} gradient flow (GF; small $\eta$), \textbf{(b)} gradient descent (GD), and \textbf{(c)} Adam. Left column: target signals (solid gray), high-dimensional RNN outputs (solid colors), and effective RNN models (dashed black). Right columns: empirical distributions of the components of the parameter vectors $\bm \theta \in \{\bm{m}, \bm{u}, \bm{v}, \bm{z}\}$ at convergence. Under GF (top), these distributions remain approximately Gaussian, and the effective model accurately captures the simulated outputs, consistent with theory. In contrast, under GD (middle) and Adam (bottom), the Gaussian assumption breaks down, and the effective theory fails to capture simulated output.}
\label{fig:9}
\end{figure}

Furthermore, unlike the linear case, the nonlinear theory relies on the components of the high-dimensional parameter vectors $\bm \theta$ remaining approximately Gaussian during training. Here, we complement the main text by (i) verifying that the weights remain approximately Gaussian under a small learning rate ($\eta = 0.05$; Fig.~\ref{fig:9}a), and (ii) showing that when this assumption breaks (e.g., with larger learning rates ($\eta=0.5$) or Adam ($\eta=0.001$)), the theory is no longer valid (Fig~\ref{fig:9}b,c).

\newpage
\subsection{History-dependent training protocol}
\label{app:B5}
\begin{figure}[!h]
\includegraphics[width=\linewidth]{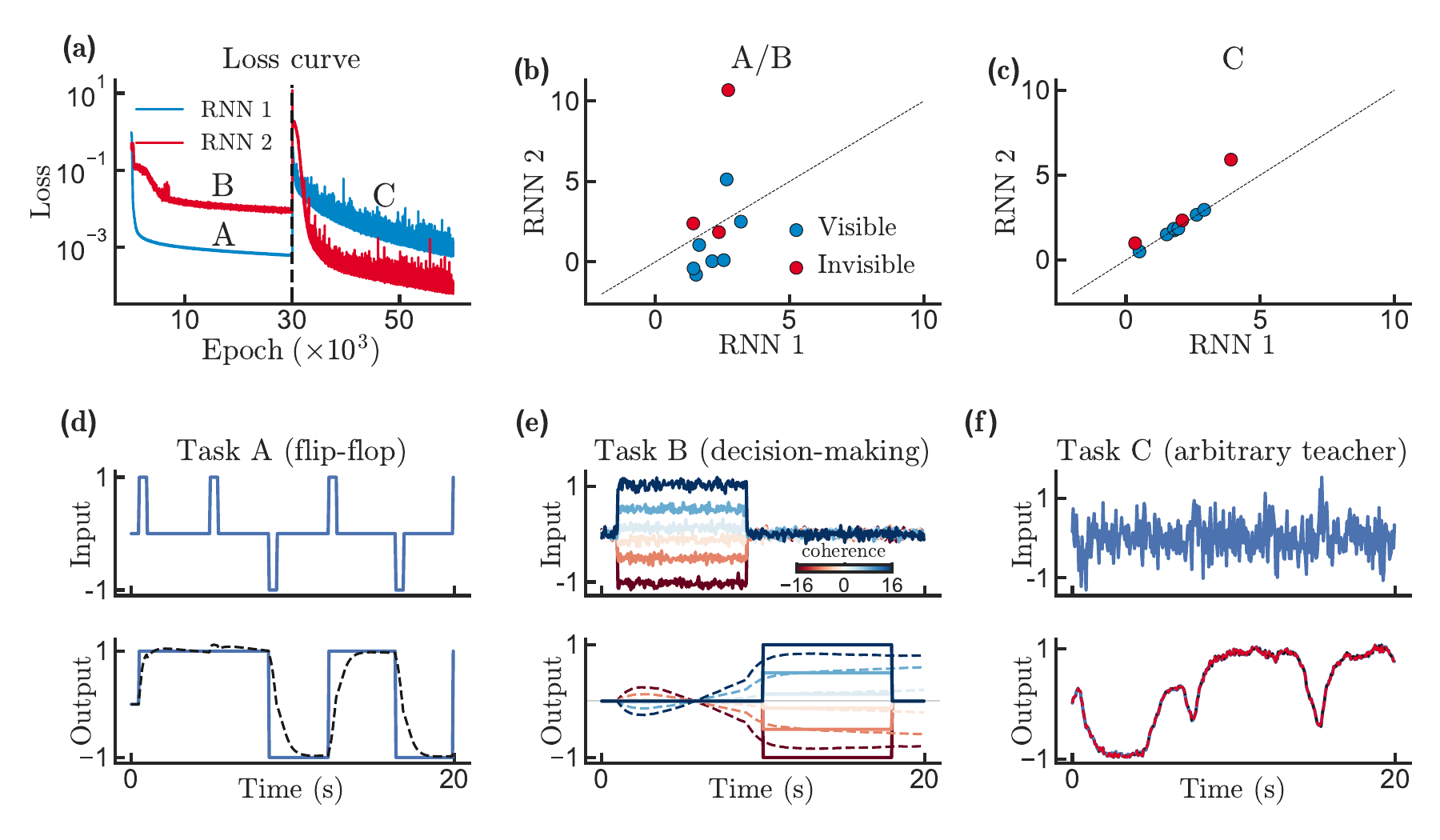}
\caption{\textbf{(a)} Training loss for the A/B $\rightarrow$ C protocol, for an example run of network 1 (A$\rightarrow$C; blue) and network 2 (B$\rightarrow$C; red). \textbf{(b)} Overlaps at the end of phase 1 (A/B; epoch 30,000), showing that both loss-visible (blue) and loss-invisible (red) overlaps settle to distinct values. \textbf{(c)} After training on task C (epoch 60,000), loss-visible overlaps converge to the same values, while loss-invisible overlaps remain distinct. \textbf{(d–f)} Example inputs (top) and target vs.\ predicted outputs (bottom) for the three tasks: (d) flip-flop, (e) stimulus integration (showing 6 different coherence levels), and (f) arbitrary teacher signal. Solid and dashed lines denote targets and network predictions, respectively.}
\label{fig:10}
\end{figure}

We consider a history-dependent training protocol consisting of three tasks. Task A is identical to the flip-flop task described above (\ref{app:B4}). Task B is a stimulus-integration decision-making task (see below). And Task C is a teacher-student task, where both networks are trained to reproduce the output of a pre-defined teacher network in response to white noise input (Fig.~\ref{fig:10}f). The teacher network overlaps are defined as
\[
\bigl(\sigma_{zm}, \sigma_{zu}, \sigma_{vm}, \sigma_{vu}, \sigma_{mu}, \|\bm m\|^2, \|\bm u\|^2\bigl)
= \bigl( 0.5,\, 2.3,\, 2.0,\, 1.5,\, 1.6,\, 1.8,\, 2.2 \bigl)
\]
Training proceeds in two phases: Phase A/B and Phase C. In Phase A/B, networks are trained on either Task A or Task B for 30{,}000 epochs, using an initial learning rate of $\eta = 0.01$. The learning rate is reduced by a factor of 5 whenever the loss falls below $0.015$. In Phase C, training continues for an additional 30{,}000 epochs with a fixed learning rate of $\eta = 0.001$. To ensure consistency with the Gaussian assumptions underlying the low-dimensional nonlinear theory, training is performed directly in the overlap space using the corresponding preconditioned $\bm G$. The initial overlap values are sampled as follows: cross-overlaps $\sigma_{zm}, \sigma_{zu}, \sigma_{vm}, \sigma_{vu}, \sigma_{mu}, \sigma_{zv}$ are drawn from $|\mathcal{N}(0, 0.4)|$, and the squared norms $\|\bm m\|^2, \|\bm u\|^2, \|\bm v\|^2, \|\bm z\|^2$ are drawn uniformly from $\mathcal{U}[0.5, 2.0]$. Within-episode dynamics are integrated with time step $\Delta t = 0.05$ over episodes of length $T=20\,\mathrm{s}$. For both phases, we used a batch size of 128. 

\paragraph{Decision-making task}
This task requires the network to integrate a noisy evidence input to produce a continuous output proportional to the stimulus strength (Fig.~\ref{fig:10}e). Each trial consists of a stimulus period of duration $t_{\mathrm{stim}}$, a brief delay $t_{\mathrm{delay}}$, and a response period. During the stimulus period, the input channel receives a noisy signal $x(t) = c + \xi(t)$, where $c$ is a coherence level chosen from a discrete set $\mathcal{C}\in\{\pm2,\pm8,\pm16\}$ and $\xi(t)\sim\mathcal{N}(0,0.05^2)$ is zero-mean Gaussian noise. During the response period the target value is defined as $y(t) = y_{\mathrm{amp}} (c/c_{\mathrm{max}})$, with $y_{\mathrm{amp}}=1.0$ and $c_{\mathrm{max}}$ is the maximum absolute coherence. The network is trained using a masked mean squared error (MSE) loss that is active only during the decision period. 

\begin{figure}[!h]
\includegraphics[width=\linewidth]{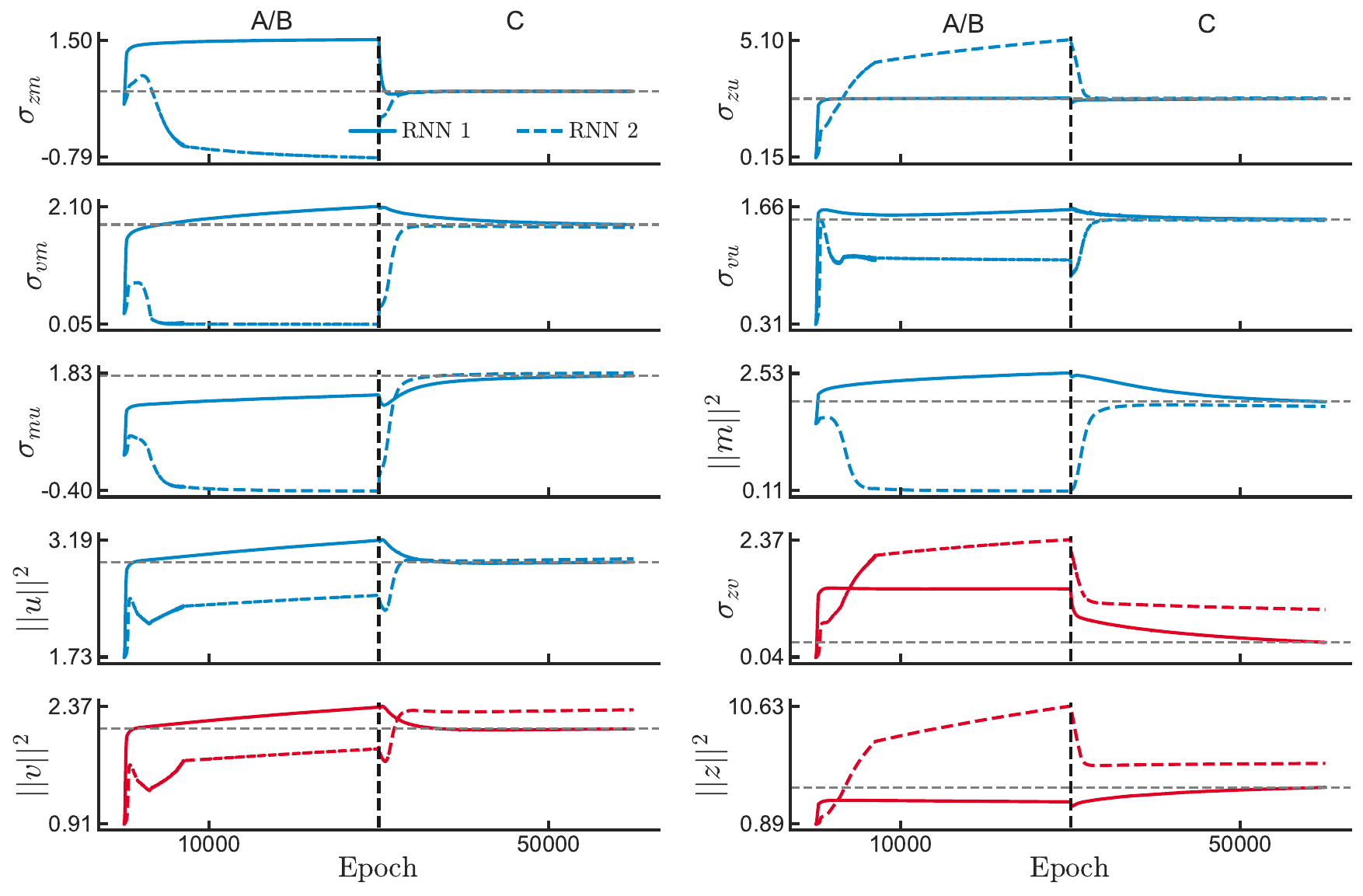}
\caption{Trajectories of all ten overlaps in the A/B $\rightarrow$ C for an example run. Blue traces denote loss-visible overlaps, red traces denote loss-invisible overlaps, with solid and dashed lines corresponding to networks 1 and 2, respectively. After retraining on task C, loss-visible overlaps converge to the same values (up to infinitesimal differences due to imperfect convergence), while loss-invisible overlaps retain distinct values, reflecting history-dependent memory. Gray dashed lines denote the converged overlap values of network 1 for Task C.}
\label{fig:11}
\end{figure}

\paragraph{Classification}
To test whether task history can be decoded from the overlaps, we trained a logistic-regression classifier (the precise choice of classifier is not critical; similar results are obtained using SVM) to classify networks first trained on Task A from those first trained on Task B. Overlap vectors were extracted from 10 independent runs (10 $\times$ 2 twins network) at three stages: initialization, after Phase A/B, and after Phase C. Classification was performed separately using either the loss-visible or loss-invisible overlaps. At each stage, the 20 samples (10 per class) were split into training and test sets (16/4), and evaluation was repeated over 50 random splits. Performance is reported as the mean and standard deviation of test accuracy across splits. Importantly, before classification, white noise was added to the overlap features to both reflect realistic variability and prevent the classifier from exploiting infinitesimal differences arising from imperfect convergence (see Fig.~\ref{fig:11}; the results are largely insensitive to the precise choice of noise level).

\clearpage
\section{Derivation of the augmented Gram matrix $\bar{\bm G}$}
\label{app:C}
To derive the augmented $\bar{\bm G}$ matrix, we define $\bar{\bm\sigma}(\bm\theta)$ as the exhaustive collection of all quadratic scalars formable from the four vectors $\{\bm m,\bm u,\bm v,\bm z\}$. This set comprises the six pairwise overlaps and the four squared norms, encompassing both loss-visible and loss-invisible quantities
\begin{equation}
\bm\theta =
\begin{bmatrix}\bm m\\ \bm u\\ \bm v\\ \bm z\end{bmatrix}\in\mathbb{R}^{4N},
\qquad
\bar{\bm\sigma}(\bm\theta)=
\begin{bmatrix}
\sigma_{zm}\\ \sigma_{zu}\\ \sigma_{vm}\\ \sigma_{vu}\\ \sigma_{mu}\\ \sigma_{zv}\\
\|\bm m\|^2\\ \|\bm u\|^2\\ \|\bm v\|^2\\ \|\bm z\|^2
\end{bmatrix}\in\mathbb{R}^{10}    
\end{equation}

The full Jacobian is given by
\begin{equation}
\bar{\bm D}(\bm\theta)=\frac{\partial \bar{\bm\sigma}}{\partial \bm\theta}
\;=\; \frac{1}{N}
\begin{bmatrix}
\bm z^\top & 0 & 0 & \bm m^\top \\
0 & \bm z^\top & 0 & \bm u^\top \\
\bm v^\top & 0 & \bm m^\top & 0 \\
0 & \bm v^\top & \bm u^\top & 0 \\
\bm u^\top & \bm m^\top & 0 & 0 \\
0 & 0 & \bm z^\top & \bm v^\top \\
2\bm m^\top & 0 & 0 & 0 \\
0 & 2\bm u^\top & 0 & 0 \\
0 & 0 & 2\bm v^\top & 0 \\
0 & 0 & 0 & 2\bm z^\top
\end{bmatrix}\in\mathbb{R}^{10\times 4N}
\end{equation}
and the associated $10\times 10$ Gram matrix, defined as $\bar{\bm G}(\bm\theta) = \bar{\bm D}(\bm\theta)\bar{\bm D}(\bm\theta)^\top$
\begin{equation}
{\normalsize
\setlength{\arraycolsep}{0pt}
\renewcommand{\arraystretch}{1.4}
\frac{1}{N}\begin{bmatrix}
\|\bm z\|^2+\|\bm m\|^2 & \sigma_{mu} & \sigma_{zv} & 0 & \sigma_{zu} & \sigma_{vm} & 2\sigma_{zm} & 0 & 0 & 2\sigma_{zm}\\
\sigma_{mu} & \|\bm z\|^2+\|\bm u\|^2 & 0 & \sigma_{zv} & \sigma_{zm} & \sigma_{vu} & 0 & 2\sigma_{zu} & 0 & 2\sigma_{zu}\\
\sigma_{zv} & 0 & \|\bm v\|^2+\|\bm m\|^2 & \sigma_{mu} & \sigma_{vu} & \sigma_{zm} & 2\sigma_{vm} & 0 & 2\sigma_{vm} & 0\\
0 & \sigma_{zv} & \sigma_{mu} & \|\bm v\|^2+\|\bm u\|^2 & \sigma_{vm} & \sigma_{zu} & 0 & 2\sigma_{vu} & 2\sigma_{vu} & 0\\
\sigma_{zu} & \sigma_{zm} & \sigma_{vu} & \sigma_{vm} & \|\bm m\|^2+\|\bm u\|^2 & 0 & 2\sigma_{mu} & 2\sigma_{mu} & 0 & 0\\
\sigma_{vm} & \sigma_{vu} & \sigma_{zm} & \sigma_{zu} & 0 & \|\bm z\|^2+\|\bm v\|^2 & 0 & 0 & 2\sigma_{zv} & 2\sigma_{zv}\\
2\sigma_{zm} & 0 & 2\sigma_{vm} & 0 & 2\sigma_{mu} & 0 & 4\|\bm m\|^2 & 0 & 0 & 0\\
0 & 2\sigma_{zu} & 0 & 2\sigma_{vu} & 2\sigma_{mu} & 0 & 0 & 4\|\bm u\|^2 & 0 & 0\\
0 & 0 & 2\sigma_{vm} & 2\sigma_{vu} & 0 & 2\sigma_{zv} & 0 & 0 & 4\|\bm v\|^2 & 0\\
2\sigma_{zm} & 2\sigma_{zu} & 0 & 0 & 0 & 2\sigma_{zv} & 0 & 0 & 0 & 4\|\bm z\|^2
\end{bmatrix}}
\label{Eq.C3}
\end{equation}

\begin{figure}[!h]
    \centering
    \includegraphics[width=\linewidth]{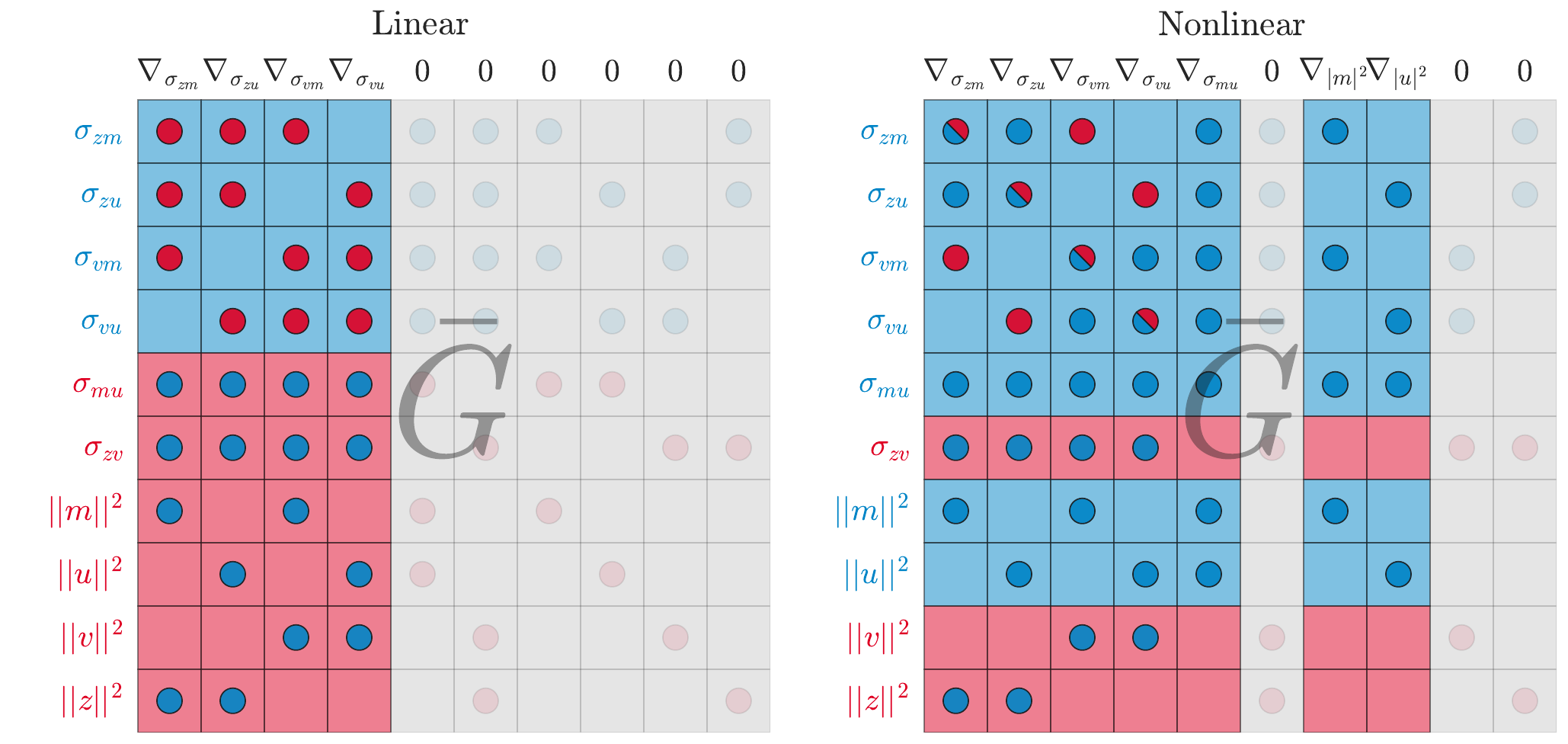}
\caption{Augmented $10\times10$ Gram matrix $\bar{\bm G}$ for the linear (left) and nonlinear (right) rank-1 RNNs. Rows correspond to overlaps being updated, and columns to the associated loss gradients. Blue letter/dots denote loss-visible overlaps, while red letter/dots denote loss-invisible overlaps. Colored circles indicate the coefficient of $\bar{\bm G}$, for which the gradient is nonzero. In the linear case, the structure is cleanly separated: updates of loss-visible overlaps depend only on the coefficients of loss-invisible quantities, and vice versa. Note that the top-left blue block of the linear matrix corresponds exactly to the $\bm G(\bm \theta)$ matrix derived in Eq.~\ref{Eq.10} of the main text.  In the nonlinear case, this separation is broken. Blue--red mixed circles highlight entries where visible and invisible quantities are coupled, implying that the learning dynamics depend jointly on both sets of overlaps. This mixing reflects the fact that quantities that are invisible in the linear model become visible in the nonlinear model.}
\label{fig:12}
\end{figure}

\medskip
Crucially, these derivations hold for any RNN where the connectivity is formed by these four vectors. While the network’s nonlinearity changes the functional form of the loss, it does not change the Jacobian $\bar{\bm{D}}$ or the Gram matrix $\bar{\bm{G}}$. The network linearity simply sets the boundary between visible and invisible overlaps, shaping how gradient flow evolves in overlap space (Fig.~\ref{fig:12}).

\clearpage
\section{Rank-2 linear RNN}
\label{app:D}
We provide a brief extension of our analysis to a rank-2 linear RNN with recurrent connectivity
\begin{equation}
    \bm{W} = \frac{1}{N} \sum_{j=1}^2 \bm{u}_j \bm{v}_j^\top
\end{equation}

The parameter set consists of six vectors $\{\bm{m}, \bm{u}_1, \bm{u}_2, \bm{v}_2, \bm{v}_2, \bm{z}\} \subset \mathbb{R}^N$. With zero initial condition $\bm{h}(0)=\bm{0}$, the dynamics remain confined to the 3-dimensional subspace $\mathrm{span}\{\bm m,\bm u_1,\bm u_2\}$. We write
\begin{equation}
\bm h(t)=\kappa_m(t)\bm m+\kappa_{u_1}(t)\bm u_1+\kappa_{u_2}(t)\bm u_2,
\qquad
\boldsymbol{\kappa}(t)=\begin{bmatrix}\kappa_m\\ \kappa_{u_1}\\ \kappa_{u_2}\end{bmatrix}
\end{equation}
The effective within-episode dynamics and readout are
\begin{equation}
\begin{aligned}
    \dot{\boldsymbol{\kappa}}(t)
    &=
    -\boldsymbol{\kappa}(t)
    +
    \begin{bmatrix} 
    0 & 0 & 0 \\ 
    \sigma_{v_1 m} & \sigma_{v_1 u_1} & \sigma_{v_1 u_2} \\ 
    \sigma_{v_2 m} & \sigma_{v_2 u_1} & \sigma_{v_2 u_2} 
    \end{bmatrix}
    \boldsymbol{\kappa}(t)
    +
    \begin{bmatrix} 1 \\ 0 \\ 0 \end{bmatrix} x(t)
    \\[8pt]
    \hat{y}(t)
    &=
    \begin{bmatrix} \sigma_{zm} & \sigma_{zu_1} & \sigma_{zu_2} \end{bmatrix}
    \boldsymbol{\kappa}(t)
\end{aligned}
\end{equation}
Thus, the input--output map, and hence the loss, is fully determined by 9 loss-visible overlaps. To derive the resulting overlap dynamics, we introduce a compact matrix formulation (alternative to that presented in the main text) that scales naturally to higher-rank networks. To this end, we collect the six parameter vectors into
\begin{equation}
\bm\Theta = \begin{bmatrix} \bm m & \bm u_1 & \bm u_2 & \bm v_1 & \bm v_2 & \bm z \end{bmatrix} \in\mathbb R^{N\times6}
\end{equation}
and define the Gram matrix $\bm Q$, which contains all 21 pairwise overlaps

\begin{equation}
\setlength{\arraycolsep}{3pt}
\bm Q = \frac{1}{N}\bm\Theta^\top\bm\Theta =
\begin{bmatrix}
\|\bm m\|^2
& \sigma_{mu_1}
& \sigma_{mu_2}
& \sigma_{v_1m}
& \sigma_{v_2m}
& \sigma_{zm}
\\
\sigma_{mu_1}
& \|\bm u_1\|^2
& \sigma_{u_1u_2}
& \sigma_{v_1u_1}
& \sigma_{v_2u_1}
& \sigma_{zu_1}
\\
\sigma_{mu_2}
& \sigma_{u_1u_2}
& \|\bm u_2\|^2
& \sigma_{v_1u_2}
& \sigma_{v_2u_2}
& \sigma_{zu_2}
\\
\sigma_{v_1m}
& \sigma_{v_1u_1}
& \sigma_{v_1u_2}
& \|\bm v_1\|^2
& \sigma_{v_1v_2}
& \sigma_{zv_1}
\\
\sigma_{v_2m}
& \sigma_{v_2u_1}
& \sigma_{v_2u_2}
& \sigma_{v_1v_2}
& \|\bm v_2\|^2
& \sigma_{zv_2}
\\
\sigma_{zm}
& \sigma_{zu_1}
& \sigma_{zu_2}
& \sigma_{zv_1}
& \sigma_{zv_2}
& \|\bm z\|^2
\end{bmatrix}
\end{equation}
Note that $\bm Q$ is distinct from the preconditioning Gram matrix $\bm G=\bm D\bm D^\top$ introduced in the main text. The nine cross-block entries of $\bm Q$ are loss-visible, whereas the remaining 12 entries are loss-invisible. Let $\nabla_{ab}=\partial\mathcal L/\partial\sigma_{ab}$, and define the symmetric gradient matrix

\begin{equation}
\setlength{\arraycolsep}{3pt}
\bm M = \begin{bmatrix}
0 & 0 & 0
& \nabla_{v_1m}
& \nabla_{v_2m}
& \nabla_{zm}
\\
0 & 0 & 0
& \nabla_{v_1u_1}
& \nabla_{v_2u_1}
& \nabla_{zu_1}
\\
0 & 0 & 0
& \nabla_{v_1u_2}
& \nabla_{v_2u_2}
& \nabla_{zu_2}
\\
\nabla_{v_1m}
& \nabla_{v_1u_1}
& \nabla_{v_1u_2}
& 0 & 0 & 0
\\
\nabla_{v_2m}
& \nabla_{v_2u_1}
& \nabla_{v_2u_2}
& 0 & 0 & 0
\\
\nabla_{zm}
& \nabla_{zu_1}
& \nabla_{zu_2}
& 0 & 0 & 0
\end{bmatrix}
\end{equation}

Because the loss depends only on the visible overlaps, gradient flow in parameter space is given by
\begin{equation}
\dot{\bm\Theta} = -\frac{1}{N}\bm\Theta\bm M
\end{equation}
Using $\bm Q=\bm\Theta^\top\bm\Theta/N$ and the product rule, we obtain

\begin{equation}
\dot{\bm Q} = \frac{1}{N} \left( \dot{\bm\Theta}^{\top}\bm\Theta + \bm\Theta^\top\dot{\bm\Theta} \right) = -\frac{1}{N} \left( \bm M\bm Q+\bm Q\bm M \right)
\end{equation}

\begin{figure}[!h]
    \centering
    \includegraphics[width=\linewidth]{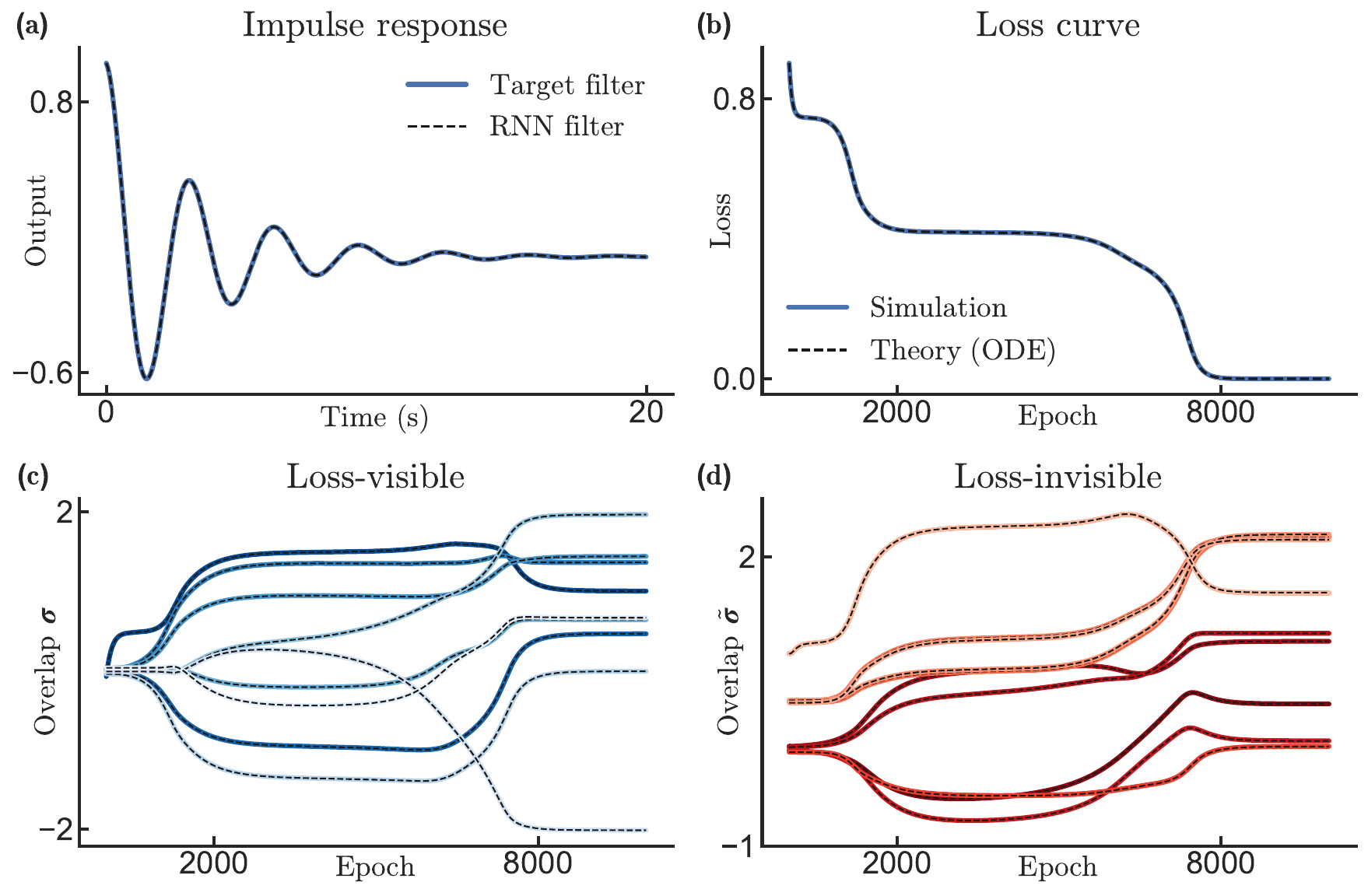}
\caption{\textbf{(a)} Impulse response of the target damped-oscillatory filter (solid blue) and the final learned RNN response (dashed black). \textbf{(b)} Training loss for the full high-dimensional simulation (solid blue) and the overlap-based ODE theory (dashed black). \textbf{(c)} Dynamics of the 9 loss-visible and \textbf{(d)} 12 invisible overlaps, comparing numerical simulations (solid) with theoretical predictions (dashed). }
\label{fig:13}
\end{figure}

This single matrix equation defines a closed $21$-dimensional system that fully characterizes learning in the rank-2 linear RNN and makes explicit how the loss-invisible entries of $\bm Q$ shape the evolution of its loss-visible entries. To validate this derivation, we train the rank-2 RNN to emulate the response of a second-order filter (damped sinusoidal~\cite{bordelon2025dynamically}):
\begin{equation}
    y^\star(t) = e^{-c^\star t} \cos(\omega^\star t)
\end{equation} 
This target filter represents an oscillatory dynamics with frequency $\omega^\star$ and damping rate $c^\star$ (set to 2 and 0.3, respectively). Aside from initializing the RNN with rank-2 connectivity, all simulation details remain identical to those used in the rank-1 analysis App.~\ref{app:A41}. Numerically training a high-dimensional RNN on this task, we find that both the loss and overlap trajectories align perfectly with the predictions of our scalar ODE system. This demonstrates the generality of our derivation beyond the rank-1 case (Fig.~\ref{fig:13}).

\paragraph{Note} For a general rank-$r$ architecture with multiple inputs $\bm m_{\text{in}}$ and outputs $\bm z_{\text{out}}$, the derivation follows the same logic: define the full set of pairwise overlaps and apply the chain rule to obtain their induced dynamics. In this general case, the total number of overlaps—both loss-visible and loss-invisible—scales as $\mathcal{O}((2r + \bm m_{\text{in}} + \bm z_{\text{out}})^2)$. While this expression grows quadratically with the rank and the number of input/output channels, it remains strictly independent of the network size $N$. In this sense, the resulting dynamics remain tractable.


\newpage
\end{document}